\documentclass[journal]{IEEEtran} 
\IEEEoverridecommandlockouts

\usepackage[utf8]{inputenc}

\usepackage{mathbbol}

\usepackage{esvect}

\usepackage[percent]{overpic}

\usepackage{tabu}

\usepackage{multicol}

\usepackage{multirow}
\def\mr{\multirow}
\def\mc{\multicolumn}

\usepackage[overload]{empheq}

\usepackage{bm}

\usepackage{xcolor}
\definecolor{dg}{rgb}{0.1, 0.6, 0.2}       
\definecolor{b}{rgb}{0.0, 0.0, 1}          

\usepackage{epsfig}
\usepackage{float}
\usepackage{color}
\usepackage{url}

\usepackage{algorithmic}
\usepackage[ruled,linesnumbered,vlined]{algorithm2e}

\usepackage{amssymb, amsfonts}
\usepackage{amsmath}
\usepackage{mathrsfs}

\usepackage{amsthm}
\usepackage{mathtools}

\let\labelindent\relax
\usepackage{enumitem}       
\setenumerate[enumerate]{align=left}

\usepackage{graphicx}
\usepackage{subcaption}

\usepackage{stackengine}

\usepackage[flushleft]{threeparttable}

\usepackage{balance}

\usepackage{soul}

\usepackage{cuted}

\usepackage{relsize}

\newcommand{\norm}[1]{\left\lVert#1\right\rVert}

\newcommand{\mypartder}[2]{\frac{\partial}{\partial #2}\ll[#1\rr]}

\makeatletter
\newlength\tmp@\newlength\t@mp
\newcommand{\comp}[3]
  {\mathop{ \settowidth\tmp@{$\displaystyle\mathop{#1}^{#3}_{#2}$}
  \hbox to \tmp@{\hss \settowidth\t@mp{$\displaystyle #1$}\setlength\t@mp{.45\t@mp}
  $\displaystyle\mathop{#1}^{\hspace\t@mp #3}_{\hspace{-\t@mp}#2}$
  \hss} }}
\makeatother

\newcommand{\Int}[2]
{\int_{#1}^{#2}}

\DeclareMathOperator*{\argmin}{argmin}

\usepackage[bookmarks=true]{hyperref}

\hypersetup{
    colorlinks=true,
    linkcolor=blue,
    filecolor=blue,      
    urlcolor=blue,
    citecolor=blue,
}

\def\g{\gamma}

\def\o{\omega}

\def\D{\Delta}
\def\S{\Sigma}

\def\R{\mathbb{R}}

\def\ll{\left}
\def\rr{\right}

\def\rot{\vbf{R}}
\def\angvel{\bs{\omega}}
\def\angacc{\vbf{\alpha}}
\def\pos{\vbf{p}}
\def\vel{\vbf{v}}
\def\acc{\vbf{a}}

\newcommand{\vbf}[1]{{\bm{\mathbf{#1}}}}

\def\bias{\mathbf{b}}

\def\SO{\mathrm{SO(3)}}

\def\SE{\mathrm{SE(3)}}

\def\Exp{\mathrm{Exp}}
\def\Log{\text{Log}}

\def\Dt{ {\D t} }
\def\dt{ {\d t} }
\def\wrt{\text{w.r.t. }}

\def\X{\mathcal{X}}
\def\Y{\mathcal{Y}}
\def\Z{\mathcal{Z}}
\def\U{\mathcal{U}}
\def\I{\mathcal{I}}
\def\V{\mathcal{V}}
\def\M{\mathcal{M}}
\def\L{\mathcal{L}}
\def\P{\mathcal{P}}

\def\T{\mathcal{T}}
\def\E{\mathcal{E}}

\def\grav{\mathbf{g}}

\newtheorem{remark}{Remark}

\def\f{f}

\def\res{\vbf{r}}
\def\tf{\vbf{T}}

\newenvironment{sqbmat}
{
\ll[\begin{smallmatrix}
}
{
\end{smallmatrix}\rr]
}

\usepackage[absolute,overlay]{textpos}
\usepackage{nccmath}    
\usepackage{multimedia} 
\usepackage{booktabs}
\usepackage{array}
\usepackage{multirow}
\usepackage{adjustbox}
\usepackage{cite}

\definecolor{mGreen}{rgb}{0.36, 0.69, 0}

\newcommand{\bs}[1]{{\boldsymbol{#1}}}

\def\bfI{\vbf{I}}
\def\bzr{\vbf{0}}

\def\dt{\Delta t}
\def\GP{\mathcal{GP}}

\def\The{\bs{\theta}}
\def\Thet{\bs{\theta}_t}
\def\Thed{{\dot{\bs{\theta}}}}
\def\Thedd{\ddot{\bs{\theta}}}
\def\Thedt{{\dot{\bs{\theta}}_t}}
\def\Theddt{\ddot{\bs{\theta}}_t}
\def\Theb{\bar{\bs{\theta}}}
\def\Then{\norm{\The}}

\def\Xi{\bs{\xi}}

\def\Xid{{\dot{\bs{\xi}}}}

\def\Xidd{{\ddot{\bs{\xi}}}}

\def\bseta{\bs{\eta}}

\def\bsrho{\bs{\rho}}
\def\bsrhod{\dot{\bsrho}}
\def\twist{\bs{\tau}}
\def\wrench{\dot{\twist}}
\def\bsnu{\bs{\nu}}

\def\bsbeta{\bs{\beta}}
\def\bsQ{\bs{Q}}
\def\bsQp{\bs{Q}'}

\def\bsu{\bs{u}}
\def\bsub{\bar{\bs{u}}}

\def\bsv{\bs{v}}

\def\Jr{\bs{J}_r}
\def\JrInv{\bs{J}_r^{-1}}
\def\SOH{\bs{H}}
\def\SOHp{\bs{H}'}
\def\SOL{\bs{L}}
\def\SOLp{\bs{L}'}

\def\SES{\bs{S}}
\def\SESp{\bs{S}'}
\def\SEL{\bs{L}}
\def\SELp{\bs{L}'}
\def\SEC{\bs{C}}
\def\SECp{\bs{C}'}

\def\SEJr{\bs{\mathcal{J}}_r}
\def\SEJrInv{\bs{\mathcal{J}}_r^{-1}}
\def\SEH{\bs{\mathcal{H}}}
\def\SEHp{\bs{\mathcal{H}}'}
\def\SEL{\bs{\mathcal{L}}}
\def\SELp{\bs{\mathcal{L}'}}

\def\usqu{u^2}
\def\ucub{u^3}
\def\utes{u^4}
\def\uqui{u^5}
\def\cosu{\cos u}
\def\sinu{\sin u}

\def\DubDu{\partder{\Theb}{\The}}
\def\DubDuCP{\bsub_\bsu}

\newcommand{\Jcb}[2]{\vbf{J}^{#1}_{#2}}
\newcommand{\partder}[2]{\frac{\partial #1}{\partial #2}}

\def\frB{\texttt{B}}
\def\frW{\texttt{W}}
\def\frL{\texttt{L}}

\def\im{\texttt{I}}
\def\ca{\texttt{C}}

\def\bsx{{\boldsymbol{x}}}
\def\bsw{{\boldsymbol{w}}}

\def\bsA{{\boldsymbol{A}}}
\def\bsB{{\boldsymbol{B}}}
\def\bsQ{{\boldsymbol{Q}}}
\def\bsF{{\boldsymbol{F}}}
\def\bsW{{\boldsymbol{W}}}

\def\srpose{\ensuremath{\SO\times\R^3}}
\def\sepose{\ensuremath{\SE}}

\begin{document}

\title{A Third-Order Gaussian Process Trajectory Representation Framework with Closed-Form Kinematics for Continuous-Time Motion Estimation}

\author{
      Thien-Minh Nguyen, \IEEEmembership{Member,~IEEE},
      Ziyu Cao, \IEEEmembership{Student Member,~IEEE},
      Kailai Li, \IEEEmembership{Member,~IEEE},
      \\
      William Talbot, \IEEEmembership{Student Member,~IEEE},
      Tongxing Jin,
      Shenghai Yuan, \IEEEmembership{Member,~IEEE},
      \\
      {Timothy D. Barfoot, \IEEEmembership{Fellow,~IEEE}}, Lihua Xie, \IEEEmembership{Fellow,~IEEE}
\thanks{Thien-Minh Nguyen, Shenghai Yuan, Tongxing Jin, and Lihua Xie are with the School of Electrical and Electronic Engineering, Nanyang Technological University, 50 Nanyang Avenue, Singapore 639798. Ziyu Cao is with the Department of Electrical Engineering, Linköping University, 58183 Linköping, Sweden. Kailai Li is with the Bernoulli Institute for Mathematics, Computer Science and Artificial Intelligence, University of Groningen, 9747 AG Groningen, the Netherlands. William Talbot is with the Robotic Systems Lab (RSL), ETH Z\"urich, 8092 Zürich, Switzerland. (e-mail: \{thienminh.nguyen, shyuan, elhxie\}@ntu.edu.sg, ziyu.cao@liu.se, kailai.li@rug.nl, wtalbot@ethz.ch). Timmothy D. Barfoot is with the Autonomous Space
Robotics Laboratory (ASRL), University of Toronto, Canada. Corresponding author: Thien-Minh Nguyen}
\thanks{
This research is supported by the National Research Foundation, Singapore, under its Medium-Sized Center for Advanced Robotics Technology Innovation (CARTIN).
}
}

\maketitle

\thispagestyle{plain}
\pagestyle{plain}

\begin{abstract}

In this paper, we propose a third-order, i.e., white-noise-on-jerk, Gaussian Process (GP) Trajectory Representation (TR) framework for continuous-time (CT) motion estimation (ME) tasks. Our framework features a unified trajectory representation that encapsulates the kinematic models of both \srpose\ and \sepose\ pose representations. This encapsulation strategy allows users to use the same implementation of measurement-based factors for either choice of pose representation, which facilitates experimentation and comparison to achieve the best model for the ME task. In addition, unique to our framework, we derive the kinematic models with the \textit{closed-form temporal derivatives of the local variable of $\SO$ and $\SE$}, which so far has only been approximated based on the Taylor expansion in the literature. Our experiments show that these kinematic models can improve the estimation accuracy in high-speed scenarios.
All analytical Jacobians of the interpolated states with respect to the support states of the trajectory representation, as well as the motion prior factors, are also provided for accelerated Gauss-Newton (GN) optimization. Our experiments demonstrate the efficacy and efficiency of the framework in various motion estimation tasks such as localization, calibration, and odometry, facilitating fast prototyping for ME researchers. We release the source code for the benefit of the community. Our project is available at \url{https://github.com/brytsknguyen/gptr}.

\end{abstract}

\IEEEpeerreviewmaketitle

\section{Introduction}

\begin{figure}
    \centering
    \includegraphics[width=0.75\linewidth]{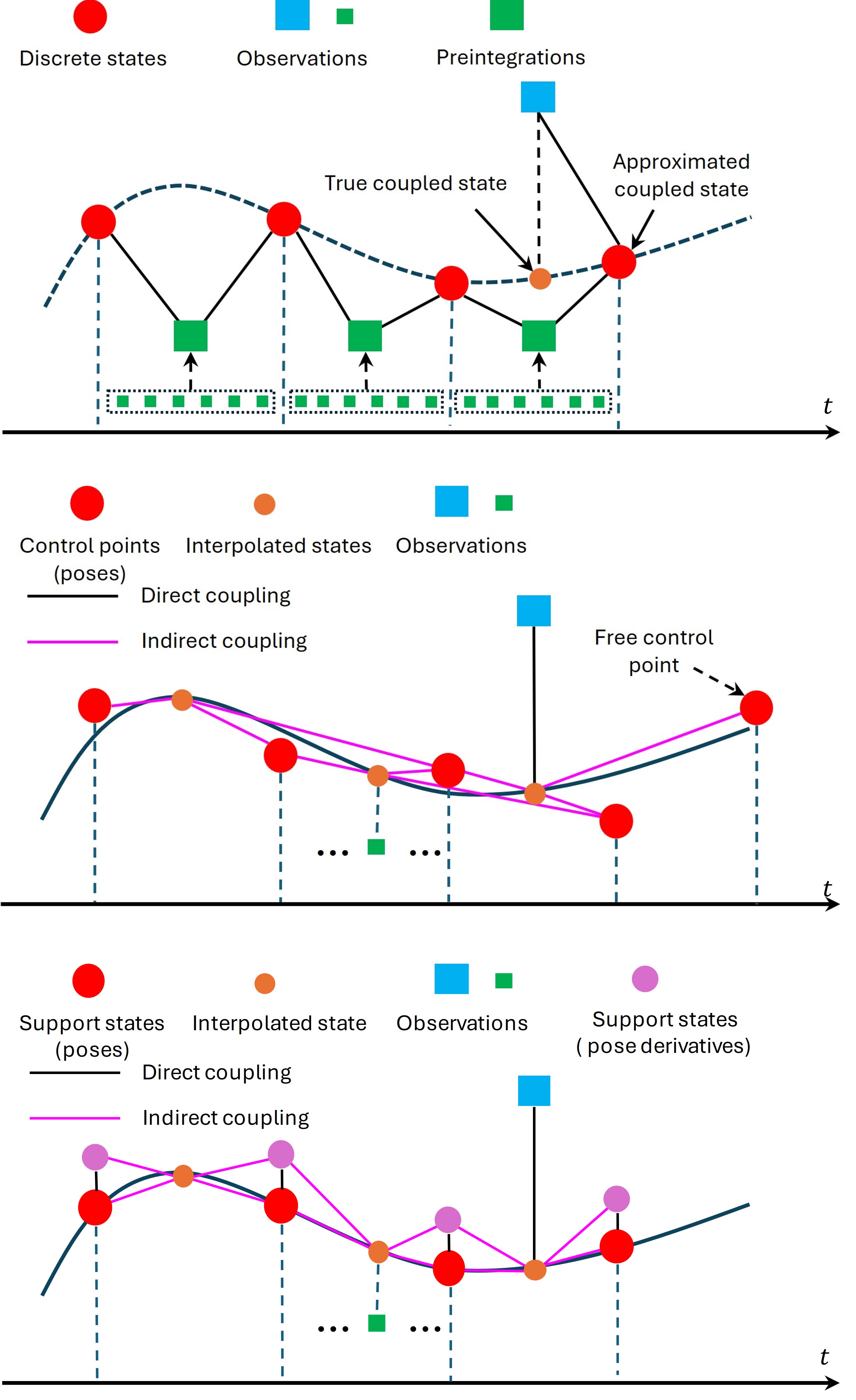}
\caption{Top: DT trajectory representation. Due to a finite number of represented states, we need to approximate, perform extra processing, or use sophisticated hardware synchronization to restrict the coupling of the observations with the state estimates.
Middle vs Bottom: Comparison of B-spline and GP-based representations of continuous time trajectory. Under the B-spline scheme, the poses and temporal derivatives can be inferred from the control points, which do not lie on the trajectory. By contrast, GP support states include trajectory poses with explicit temporal derivatives according to a chosen prior motion model.}
\label{fig: trajectory representation}
\vspace{-0.3cm}
\end{figure}

{Continuous-time (CT) Motion Estimation (ME) is a powerful technique} that rectifies several disadvantages of the conventional discrete-time (DT) approach, the biggest of which is that it restricts the coupling of observations with a few chosen discrete states, leading to some undesirable compromises (Fig. \ref{fig: trajectory representation}). On the other hand, continuous-time estimation approaches allow one to directly fuse sensor measurements at the sample acquisition time. This is very useful for high-frequency and temporally spread-out sensors such as IMU, event camera, LiDAR, etc. Besides, continuous-time trajectory representation also naturally allows the estimation of temporal offsets and querying the state at any time instance on the trajectory. {We refer to \cite{cioffi2022continuous} for a more comprehensive comparative study of CT and DT approaches for SLAM}.

{Despite its advantages, the CT paradigm remains ``underutilized" according to a recent survey \cite{talbot2024continuous}}. Several CT approaches have been proposed over the years. The simplest of which is the linear-interpolation (LI) representation that assumes a constant rate of change between two consecutive poses \cite{bosse2009continuous, bosse2012zebedee, dellenbach2022ct, nguyen2021ntuviral, nguyen2023slict}. However, the LI representation cannot represent derivative states, hence, they could not fuse IMU data directly at their timestamp.
Hence, more general approaches are needed to address this issue.

One approach to address this issue is by using a smooth polynomial representation of poses, as in B-splines methods, which allows representing the derivative states by the polynomials of lower orders. B-spline methods have been extensively applied in the batch-optimization problems for multi-sensor calibration \cite{furgale2012continuous, furgale2013unified, rehder2016extending}. 
{In recent years, B-spline has also been used for sliding-window-based CTME of mono/stereo visual(-inertial) odometry (VO / VIO) \cite{oth2013rolling, patron2015spline, mueggler2018continuous, lang2022ctrl, hug2020hyperslam, hug2022continuous, hug2025hyperion}}, UWB-IMU \cite{li2023continuous}, and LiDAR(-inertial) odometry (LO / LIO) \cite{quenzel2021real, lv2021clins, nguyen2024eigen, park2021elasticity}. However, we find that B-spline representation still has some limitations. First, the B-spline model dictates that the trajectory follows a polynomial constraint. Indeed, in \textit{motion planning}, one can choose a polynomial as the trajectory, but when estimating the motion, we cannot assume that the trajectory must follow a specific model. A simple counter-example is the velocity of a falling object under air drag, $\vel(t)= \vel_\text{terminal}\times \text{tanh}(\grav \cdot t / \vel_\text{terminal})$, which follows an exponential trajectory, instead of the polynomial. Second, because of the polynomial kinematic constraints, the control points are not actual states on the trajectory, which makes the initialization of new control points non-trivial. Finally, the B-spline also requires several control points that extend beyond the end time of the trajectory, which can move freely and make the regression problem ill-posed.

{In parallel to the development of B-spline, there has also been an active line of research on CTME based on temporal Gaussian Process (GP), which -- though not exhaustively --can be traced through a series of works \cite{tong2013gaussian, barfoot2014batch, anderson2015batch, anderson2015full, barfoot2024state}. Specifically, in the first work \cite{tong2013gaussian}, Tong et al. adapted the concept of \textit{GP regression} in machine learning literature \cite{seeger2004gaussian} to a batch simultaneous localization and mapping (SLAM) problem by using Gauss-Newton (GN) optimization. They formulated the trajectory as a composition of (possibly infinitely) many temporal basis functions, then exploited the \textit{kernel trick} in the inverse manner to ensure computational tractability. Then, in \cite{barfoot2014batch, anderson2015batch}, Barfoot and Anderson proved that the GP priors based on a class of linear-time-varying (LTV) stochastic diferential equations (SDE) driven by white noise have exactly
sparse inverse kernel matrices, resulting in solution times that are linear in the number of estimation times. Furthermore, they show that interpolation is an $O(1)$ operation. In their next work, \cite{anderson2015full}, Anderson et al., for the first time demonstrated how GP-based CTME can be extended to manifold states, hence enabling the adoption of this technique in a wide class of robotics problems, which are called Simultaneous Trajectory Estimation and Mapping (STEAM). A comprehensive monograph on GP, as well as state estimation in general, can be found at \cite{barfoot2024state}.}

In our opinion, GP-based trajectory representation offers some advantages compared to B-spline. Rather than subjectively assuming some type of kinematics, sparse GP methods often use a well-grounded kinematic model as the prior, and only assume the derivative of the highest order to be a Gaussian-distributed stochastic term. Based on the order of the derivative, we can refer to the GP model as \textit{random velocity}, \textit{random acceleration}, \textit{random jerk}, also known as white-noise-on-velocity/acceleration/jerk, or WNOV, WNOA, WNOJ. As illustrated in Fig. \ref{fig: trajectory representation}, while B-spline encodes the information on velocity and acceleration via a sequence of \textit{control points} under the interpolation rule, GP directly includes velocity and acceleration quantities in the \textit{support states}.
While retaining all the benefits of the B-spline approach, GP representation also avoids the issue of poorly-constrained control points. In addition, as the GP support states are true states on the trajectory, initialization of new support states can be done easily with the state transition model. Moreover, the state interpolation process only involves the two adjacent support states, making the organization, query, and extension of the trajectory much more straightforward than in B-spline representation. Another advantage of the GP-based methods is that the covariance can also be propagated and interpolated just like the states. {We refer readers to \cite{johnson2024continuous} for a more comprehensive comparison between GP and B-spline, whereas in this paper we will focus on the difference between GP-based approaches.}



{Similar to CT methods in general, we believe the GP-based paradigm has been underutilized despite a mature theoretical foundation, which we suspect is due to the scarcity of open-sourced and instructional works compared to DT approaches.
The most prominent existing open-sourced work to be mentioned is STEAM\footnote{\url{https://github.com/utiasASRL/steam}} \cite{barfoot2014associating}.
To the best of our knowledge, only STEAM\_ICP \cite{wu2022picking, burnett2024continuous, burnett2024imu} from the same group has followed up with an open-sourced application.
Another example of open-source GP representation is ESGPR \cite{pervsic2021spatiotemporal}. However, ESGPR only considers the calibration applications, where a 9DOF vector-space GP is used to model the linear position, velocity, and acceleration states of a target.
We also noted GPMP \cite{mukadam2018continuous} and its applications \cite{yan2017incremental, mukadam2017simultaneous, warren2018towards}; however, these works focus on motion planning instead of motion estimation.}

{
By learning from previous works, we are motivated to develop a new software framework to boost the popularity of GP-based CTME methods, named GPTR. In a sense, our work follows the spirit of STEAM with some new developments. Several important features are targeted in the design of GPTR.
Among this, the choice of pose representation is a cornerstone of the framework since this choice depends upon how forces and torques act upon the object undergoing motion~\cite{ovren2019trajectory, haarbach2018survey} and can affect the performance of the ME scheme.
However, in practice, one may not be entirely certain which pose representation is the most suitable for their application.
To mitigate this issue, we propose a unified trajectory representation for \srpose\ and \sepose\ that encapsulates the internal kinematics that are specific to each pose representation. This unified framework allows the implementation of the cost factors to stay the same regardless of the chosen pose representation, thus saving researchers the time to experiment with different models.

For the time interval between support states, which is called the \textit{motion prior interval} (MPI) and denoted as $\Dt \in \R^+$, we opt to space them uniformly and let it be a user-defined hyperparameter. This is a common choice in the literature, although strategies for non-uniform spacing could be investigated in future work, as has been done for B-splines~\cite{talbot2024continuous}.

Next, we choose a random jerk model (WNOJ) for the interpolation and motion model prior. This is a 3rd-order model which is useful in a practical sense, since it allows the direct inclusion of accelerometer measurements.
The hyperparameters of this model, the noise characteristics $\bs{\S}_\bs{\nu}, \bs{\S}_\bs{\g} \in \R^{3\times3}$, are user-defined, since they vary by experiment. Some previous works have partially explored the 3rd-order model, such as a hybrid $\SO$-WNOA + $\R^3$-WNOJ \cite{zheng2025traj}, approximated $\SE$-WNOJ \cite{tang2019white}, our work stands out by the investigation of both the \srpose-WNOJ and $\SE$-WNOJ models with closed-form kinematics under a unified trajectory representation. Experiment results show that the closed-form kinematic model can achieve higher accuracy in high-speed scenarios than the closed-form model. Moreover, in one specific scenario involving visual-inertial factor, we can observe faster convergence thanks to the closed-form kinematic model requiring fewer iterations than the approximated kinematic model.
}

Besides the mathematical structure, we also pay attention to the practical side of our framework. Specifically, our library is contained within a few C++ header files. We also integrate our framework with other popular libraries in robotics, such as Sophus\footnote{\url{https://github.com/strasdat/Sophus}},  Eigen\footnote{{\url{https://eigen.tuxfamily.org/}}} for manifold class abstraction, and Ceres Solver \cite{ceres-solver} for GN optimization, thus making the framework versatile for adoption into other projects. We also provide some minimal working examples of CTME with GPTR in various Maximum Likelihood Estimation (MLE) and Maximum A Posteriori (MAP) optimization schemes. These examples involve LiDAR, IMU, UWB, and visual factors with analytical Jacobians that follow the standard of the Ceres Solver, and are packaged under both ROS1 and ROS2 frameworks. These practical implementation approaches have been carefully deliberated to ensure ease of adoption by the community.



The main contributions of our work can be stated as follows:
\begin{itemize}

    \item {We propose a unified trajectory representation that encapsulates the internal kinematics of \srpose\ and \sepose\ to mitigate the difficulty of having to re-implement the ME scheme for each representation, thus facilitating experimentation and comparison to achieve the best model for the ME task.}

    \item {We provide the closed-form kinematics of third-order on-manifold GPs based on both $\SO\times \R^3$ and $\SE$ pose representations.}
        
    \item {We compare the performance of the proposed closed-form kinematics model with the existing approximated model in various scenarios, demonstrating the benefits of the closed-form model.}
    
    \item We release our code base, named GPTR, consisting of our GP-based CTME framework and several working examples of MLE and MAP optimization problems with visual, UWB, IMU, and LiDAR factors. These examples can serve as prototypes to boost the applications of CTME and GP methods.
\end{itemize}

The remainder of this paper is organized as follows: Sec. \ref{sec: gp on vector space} recalls the basic theory of motion estimation based on the Gaussian Process, i.e., propagation and interpolation on vector space. {Sec. \ref{sec: gp on manifold} provides the detailed derivation of closed-form kinematics of GPs on both \srpose\ and $\SE$ manifolds. Sec. \ref{sec: trajectory representation} establishes a trajectory representation that encapsulate different kinematics of \srpose\ and \sepose\ pose representations.}
Sec. \ref{sec: scenarios} details several MLE and MAP optimization schemes leveraging image, IMU, UWB, and LiDAR measurements; Sec. \ref{sec: experiment} presents the experiment results that showcase the accuracy and computation loads of the new closed-form equations, as well as the importance of selecting the pose representations. Finally, Sec. \ref{sec: conclusion} concludes the paper.

\section{Preliminaries} \label{sec: gp on vector space}

\subsection{Notations}
In this paper, a matrix is denoted by a bold uppercase letter, and vectors by a bold lowercase letter. For a matrix $\bm{A}$ (and similarly, for vectors), $\bsA^\top$ denotes its transpose. Given $\bm{A} \in \R^{m\times n}$, we also write $\bsA = [a_{ij}]$, where $a_{ij}$ refers to the entry at $i$-th row and $j$-th column, where $i = 0, \dots m-1$ and $j = 0, \dots n-1$. For two matrices $\bsA \in \R^{a\times b}$ and $\bsB \in \R^{c\times d}$, $\bsA \otimes \bsB \in \R^{ac \times bd}$ denotes their Kronecker product. 
For a number of square matrices $\{\bsA_i\}$ (including scalars), $\text{diag}(\dots \bsA_i \dots)$ denotes the block-diagonal matrix constructed from $\{\bsA_i\}$. Similarly, for $\{\bsA_i \in \R^3\}$, we denote $\text{vstack}(\dots \bsA_i \dots) \triangleq [\dots \bsA_i^\top \dots]^\top$ or $(\dots \bsA_i \dots)$ when the meaning is clear from the context.

To avoid ambiguity, we may attach a left superscript to indicate the \textit{parent} coordinate frame, and a left subscript for the \textit{child} frame. Thus, ${}^\frW_\frB\rot$ denotes the orientation of the frame $\frB$ (body) \wrt the frame $\frW$ (world). Moreover, for a vector $\vbf{v}$, ${}^\frB\vbf{v}$, ${}^\frW\vbf{v}$ explicitly refer to the coordinates of $\vbf{v}$ in the coordinate frame $\frB$ and $\frW$, respectively, and ${}^\frW\vbf{v} = {}^\frW_\frB\rot{}^\frB\vbf{v}$.
{Similarly, we define a transformation matrix ${}^\frW_\frB\tf = \ll[\begin{smallmatrix}
{}^\frW_\frB\rot &{}^\frW_\frB\pos\\ \textbf{0} &1\end{smallmatrix}\rr] \in \SE$, where $\rot \in \SO$ and $\pos \in \R^3$.} The coordinate transform of a point $\vbf{x}$ is ${}^\frW\vbf{x} = {}^\frW_\frB\rot {}^\frB \vbf{x} + {}^\frW_\frB\pos$.
When the frames are variable, to simplify notations, we may supplant the full notation of the frame with the index or time stamp. For example ${}^\frB_{\texttt{S}_i} \tf = {}_{i} \tf$ can denote the \textit{extrinsics} of the sensor $\texttt{S}_i$ in the body frame.
For a vector $\vbf{x} \in \R^3$, $\vbf{x}^\wedge$ denotes the corresponding skew-symmetric matrix. {We overload the functions $\Exp$ and $\Log$ for the mappings between $\SO$, $\SE$ with their vector tangents, i.e. $\Exp(\cdot) : \R^n \to \M$ and $\Log(\cdot) : \M \to \R^n$, $n = 3$ if $\M = \SO$ and $n=6$ if $\M = \SE$. In addition, we follow \cite{sola2018micro} for the generalization of Jacobian from vector space to Jacobian on manifold, based on the right convention for plus/minus operations, as explained in Sec. E in \cite{sola2018micro}.}

{
We use $t$ to denote the time. $t$ can be included as an argument of a time-varying variable $\bsx$, i.e. $\bsx\bs(t)$, or a subscript $\bsx_t$, or omitted when the context is clear.
}
$\mathcal{N}(\bm{\mu}, \bm{\S})$ denotes a multivariate normal distribution with mean $\bm{\mu}$, covariance $\bm{\S}$, and $\GP(\bs{\mu}(t), \bs{\kappa}(t))$ denotes a Gaussian Process with the mean function $\bs{\mu}(t)$ and the covariance function $\bs{\kappa}(t)$.

\subsection{GP Propagation}
Denote $\bsx(t)$ as a state vector of interest. Let us assume that $\bsx(t)$ evolves by the following LTV-SDE:
\begin{align}
    \dot{\bsx}(t) = \bsA(t)\bsx(t)+\bsB(t)\bseta(t),
    \ 
    \bseta(t) \sim \mathcal{N}(\bzr, \bs{\S}). \label{eq: gp ltv}
\end{align}
Also, we define $\bs{\Phi}(t, t_0)$ as the \textit{state transition matrix} of the deterministic system:
\begin{equation}
    \dot{\bsx}(t) = \bsA(t)\bsx(t),\ \bsx(t_0) = \bsx_0.
\end{equation}

From \eqref{eq: gp ltv} we have $\bsx(t) \sim \GP(\bs{\mu}(t), \bs{\kappa}(t, t_0))$ with the \textit{mean function} $\bs{\mu}(t)$ and \textit{covariance function} $\bs{\kappa}(t,t_0)$ defined as:

\begin{align} \label{eq: propogation}
    \bs{\mu}(t)
    &= \bs{\Phi}(t, t_0)\mu(t_0)
    \nonumber \\
    \bs{\kappa}(t, t_0)
    &= \bs{\Phi}(t, t_0)\bs{\kappa}(t_0, t_0)\bs{\Phi}(t, t_0)^\top + \bsQ(t, t_0)
    \nonumber \\
    \bsQ(t, t_0)
    &= \int_{t_0}^t\bs{\Phi}(t, s)\bsB(s)\bs{\S}\bsB(s)^\top\bs{\Phi}(t, s)^\top ds.
\end{align}

\subsection{GP Interpolation}
Given the samples of $\bs{\mu}(t)$ at different times $\{\bs{\mu}_k = \bs{\mu}(t_k)\}_{k=0,1,\dots,K}$, and assume that $\bsx(t)$ evolves according to \eqref{eq: gp ltv},
Barfoot et al. \cite{barfoot2014batch, barfoot2024state},
show that the mean state at time $\tau$, for $\tau \in [t_k, t_{k+1})$, can be found by:
\begin{equation} \label{eq: interpolation}
    \bs{\mu}(\tau) = \bs{\Lambda}(\tau)\bs{\mu}_k + \bs{\Psi}(\tau)\bs{\mu}_{k+1},
\end{equation}
where $\bs{\Lambda}(\tau)$ and $\bs{\Psi}(\tau)$ are defined as:
\begin{align}
    &\bs{\Lambda}(\tau) = \bs{\Phi}(\tau, t_k) - \bs{\Psi}(\tau)\bs{\Phi}(t_{k+1}, t_k),
    \\
    &\bs{\Psi}(\tau) = \bsQ(\tau, t_k)\bs{\Phi}(t_{k+1},\tau)^\top \bsQ(t_{k+1}, t_k)^{-1}.
\end{align}

\subsection{GP on Vector Space}

Given $\bs{\nu}(t) = \left(\pos(t), \vel(t), \acc(t)\right)$, where $\pos, \vel, \acc$ are the position, velocity, acceleration, and assume a random \textit{jerk} model, we have a GP based on the following linear time-invariant (LTI) system:
\begin{align} \label{eq: pva process}
    &\dot{\bs{\nu}}(t)
    = 
    \bsA
    \bs{\nu}(t)
    +
    \bsB
    \bseta_\bs{\nu}(t),
    \ 
    \bseta_\bs{\nu}(t) \sim \mathcal{N}(\bzr, \bs{\S}_\bs{\nu}),
\end{align}
where
$\bsA \triangleq
\begin{bmatrix}
    \textbf{0} &\textbf{I} &\textbf{0}\\
    \textbf{0} &\textbf{0} &\textbf{I}\\
    \textbf{0} &\textbf{0} &\textbf{0}
\end{bmatrix}$,
$
\bsB \triangleq
\begin{bmatrix}
    \textbf{0}
    \\
    \textbf{0}
    \\
    \textbf{I}
\end{bmatrix}$.

Let us define the transition matrix and the covariance matrix as follows:
\begin{align} \label{eq: F and Q from phi and integration}
    &\bsF(\Dt) \triangleq \bs{\Phi}(\Dt, 0) = e^{\bsA\dt},
    \nonumber\\
    &\bsQ_\bs{\nu}(\Dt, \bs{\S}_\bs{\nu}) \triangleq \int_{0}^\Dt\bsF(s)\bsB\bs{\S}_\bs{\nu}\bsB^\top\bsF(s)^\top ds.
\end{align}
Hence, we can obtain the discrete model:
\begin{align}
    &\bs{\nu}_{k+1} = \bsF(\Dt) \bs{\nu}_{k} + \bs{\eta}_k \triangleq \ll[\bar{\bsF}(\Dt) \otimes \textbf{I} \rr] \bs{\nu}_{k} + \bs{\eta}_k,
    \nonumber\\
    &\bs{\eta}_k \sim \mathcal{N}(\textbf{0}, \bsQ_\bsnu) \triangleq \mathcal{N}(\textbf{0}, \bar{\bsQ}_\bs{\nu}(\Dt)\otimes \bs{\S}_\bs{\nu}), \label{eq: state transition model}
\end{align}
where $\bar{\bsF}(\Dt) $ and $\bar{\bsQ}_\bs{\nu}(\Dt)$ can be quickly calculated by the following formulas:
\begin{align} \label{eq: FQ formulas}
    &\bar{\bsF}(\Dt) = \left[f_{nm}\right],\ f_{nm} =
    \begin{cases}
        \Dt^m/m!,\ & \text{if}\ m \geq n \\
        0,                 & \text{otherwise}
    \end{cases}
    ,
    \nonumber\\
    &\bar{\bsQ}_\bs{\nu}(\Dt)
    = \left[q_{nm}\right],
    \nonumber\\
    &q_{nm} = \frac{\Dt^{2D+1-n-m}}{(2D+1-n-m)(D-n)!(D-m)!},
    \nonumber\\
    &D = N-1;\ n, m \in \{0, \dots N-1\}, \Dt = t - t_0.
\end{align}
where $N$ is the order of the system, which, for the specific case of $N=3$, we have $\bar{\bsF}$ and $\bar{\bsQ}$ as:
\begin{align} \label{eq: Fs Qs explicit}
    \bar{\bsF}(\Dt)
    &=
    \begin{bmatrix}
    1       &\Dt    &0.5\Dt^2\\\
    0       &1    &\Dt\\
    0       &0    &1
    \end{bmatrix},
    \\
    \bar{\bsQ}(\Dt)
    &= 
    \begin{bmatrix}
    {\Dt^5/20} &{\Dt^4/8}  &{\Dt^3/6}\\
    {\Dt^4/8}  &{\Dt^3/3}  &{\Dt^2/2}\\
    {\Dt^3/6}  &{\Dt^2/2}  &\Dt\\
    \end{bmatrix},
\end{align}

\begin{remark}
    The explicit forms of $\bar{\bsF}$ and $\bar{\bsQ}$ for models of {second and third orders} have been frequently presented in the existing literature on GP \cite{barfoot2014batch, tang2019white, pervsic2021spatiotemporal, johnson2024continuous, zheng2025traj}. In this paper, we stipulate the formulas \eqref{eq: FQ formulas} that can generate $\bar{\bsF}$, $\bar{\bsQ}$ to {both second and third order models. These formulas are directly derived from the theoretical foundation of \cite{barfoot2014batch, tang2019white} for WNOA and WNOJ model, but expressed as implementation-friendly functions of $\Dt$ and $N > 1$ and the indices of the matrix entries, thus waiving the need of having to calculate the matrix exponential and integral \eqref{eq: F and Q from phi and integration} for every $N$.}
\end{remark}

\section{Gaussian Process on Manifolds} \label{sec: gp on manifold}

{In the previous section, we have reviewed the basics of GPs on vector space. However, in most applications, the trajectory must involve manifold states with significantly more complex kinematics. Thus, in this section, we shall detail the formulation of the third-order GP on two special manifolds of particular interest to motion estimation.

}

\subsection{GP With \texorpdfstring{\srpose}{SO3} Pose Representation} \label{sec: gptr on SO3}

\begin{table*}

    \centering
    \renewcommand{\arraystretch}{1.1}
    \normalsize
    \begin{threeparttable}
    \caption{Derivations of the kinematic models for GP on $\SO$ manifold} \label{tab: kinematic model derivation so3}
    \begin{tabular}{l|l}
    \hline

    \multicolumn{1}{c}{\textbf{Approximated Approach} \cite{tang2019white, johnson2024continuous}}
    &\multicolumn{1}{c}{\textbf{Closed-form Approach (Ours)}}

    \\\hline
    \multicolumn{2}{c}{\textbf{Global} $\to$ \textbf{Local}:}
    \\

    \multicolumn{2}{c}
    {
    \begin{minipage}{0.425\linewidth}
      \begin{equation}
        \The = \Log(\rot_k^{-1}\rot)
      \end{equation}
    \end{minipage}
    }
    \\

    \multicolumn{2}{c}
    {
    \begin{minipage}{0.425\linewidth}
    \begin{equation}
        \Thed = \JrInv(\The)\angvel,
    \end{equation}
    \end{minipage}
    }
    \\

    \multicolumn{1}{l|}
    {
    \begin{minipage}{0.425\linewidth}
    \vspace{0.1cm}
    \begin{align}
        \Thedd
        =
        \frac{d\Thed}{dt} = \JrInv(\The)\frac{d\angvel}{dt}
        +
        \frac{d}{dt}\left[\JrInv(\The)\right]
        \angvel
        \nonumber\\
        \approx
        \JrInv(\The)\angacc
        +
        \frac{d}{dt}\left[\bfI + \frac{1}{2}\The^\wedge\right]\angvel
        \nonumber\\
        =
        \JrInv(\The)\angacc 
        - \frac{1}{2}\angvel^\wedge\Thed
        \qquad\qquad\mspace{5mu}
        \label{eq: local to global mappings approx}
    \end{align}
    \vspace{0.05cm}
    \end{minipage}
    }
    
    &
    
    \multicolumn{1}{l}
    {
    \begin{minipage}{0.425\linewidth}
    \begin{align}
        \Thedd
        =
        \frac{d\Thed}{dt} = \JrInv(\The)\frac{d\angvel}{dt}
        +
        \frac{\partial }{\partial\The}\left[\JrInv(\The)\angvel\right]
        \frac{d\The}{dt},
        \nonumber\\
        \triangleq
        \JrInv(\The)\angacc + \SOHp_1(\The, \angvel)\Thed,
        \qquad\qquad
        \label{eq: local to global mappings}
        \\
        \SOHp_1(\bsu, \bsv) = \partder{\JrInv(\bsu)\bsv}{\bsu}
        \qquad\qquad\qquad\qquad\quad
    \end{align}
    \end{minipage}
    }

    \\\hline
    \multicolumn{2}{c}{\textbf{Global} $\xleftarrow{}$ \textbf{Local}:}
    \\

    \multicolumn{2}{c}
    {
    \begin{minipage}{0.425\linewidth}
      \begin{equation}
        \rot = \rot_k\Exp(\The)
        \label{eq: 1}
      \end{equation}
    \end{minipage}
    }
    \\

    \multicolumn{2}{c}
    {
    \begin{minipage}{0.425\linewidth}
    \vspace{0.1cm}
    \begin{equation}
        \angvel
        =
        \frac{\partial }{\partial\The}\left[\rot_k\Exp(\The)\right]
        \frac{d\The}{dt}
        = \Jr(\The)\Thed
        \label{eq: 2}
    \end{equation}
    \vspace{0.01cm}
    \end{minipage}
    }
    \\

    \multicolumn{1}{l|}
    {
    \begin{minipage}{0.425\linewidth}
    \begin{align}
        \angacc
        \overset{\eqref{eq: local to global mappings approx}}{\approx}
        \Jr(\The)\Thedd
        + \frac{1}{2}\Jr(\The)\angvel^\wedge\Thed
        \qquad\quad\mspace{5mu}
    \end{align}
    \end{minipage}
    }
    
    &
    
    \multicolumn{1}{l}
    {
    \begin{minipage}{0.425\linewidth}
    \vspace{0.1cm}
    \begin{align}
        \angacc
        =
        \frac{d\angvel}{dt}
        = \Jr(\The)\frac{d\Thed}{dt}
        + \frac{\partial }{\partial\The}
          \left[\Jr(\The)\Thed\right]
          \frac{d\The}{dt}
        \nonumber
        \\
        \triangleq \Jr(\The)\Thedd + \SOH_1(\The, \Thed)\Thed,
        \qquad\quad
        \label{eq: glocal to local mappings}
        \\
        \SOH_1(\bsu, \bsv) \triangleq \partder{\Jr(\bsu)\bsv}{\bsu}
        \qquad\qquad\qquad\qquad
    \end{align}
    \vspace{0.05cm}
    \end{minipage}
    }
    
    \vspace{0.1cm}
    \\
    \hline
    \end{tabular}

    \begin{tablenotes}
        \small
        \item Note: $\Jr$ is the \textit{right Jacobian of} $\SO$ and $\JrInv$ is its inverse \cite{sola2018micro}. In Appendix \ref{app: Jr JrInv H Hprime}, we detail the procedure to obtain the closed forms of the mappings $\Jr$, $\JrInv$, $\SOH_1(\cdot, \cdot)$, $\SOHp_1(\cdot, \cdot)$. These procedures will also be used as illustrations for a similar process in the \sepose\ case.
    \end{tablenotes}
    \end{threeparttable}

\end{table*}

{Also called the split representation, in this formulation, two GPs are interpolated and propagated separately. In \cite{zheng2025traj}, a composition of a second-order GP for $\SO$ and a third-order GP for $\R^3$ was proposed. In this section, we will detail the formulation of the third-order GP of $\SO$, which will also provide some foundation for the $\SE$ GP.}

Let us define the state $\ll(\rot_t,\ \angvel_t,\ \angacc_t\rr) \in \SO \times \R^3 \times \R^3$, where $\rot_t$, $\angvel_t$, $\angacc_t$ are respectively the \textit{rotation}, \textit{angular velocity}, and \textit{angular acceleration} states.
The evolution of $\ll(\rot_t,\ \angvel_t,\ \angacc_t\rr)$ can be modeled by the following process:
\begin{align} \label{eq: rdot}
    \dot{\rot}_t = \rot_t\angvel_t^\wedge,
    \ 
    \dot\angvel_t = \angacc_t,
    \ 
    \dot{\angacc}_t \sim \mathcal{N}(\bzr, \bs{\S}).
\end{align}
However, \eqref{eq: rdot} is nonlinear, and no closed-form solution of the mean and covariance of this process is known to the best of our knowledge.
To resolve this issue, for each interval $[t_k, t_{k+1})$, Anderson et al. \cite{anderson2015full} defined the \textit{local states} $\The_t$ and $\bs{\gamma}_t$ as follows:
\begin{align}
    \The_t
    &\triangleq
    \text{Log}(\rot_{k}^{-1}\rot_t) \in \R^3,
    \ 
    \bs{\gamma}_t
    =
    (\The_t, \Thed_t, \Thedd_t) \in \R^9, \label{eq: def gamma}
\end{align}
which belong to a vector space, and we can assume $\bs{\gamma}_t$ evolves by the same system as \eqref{eq: pva process}:
\begin{align} \label{eq: gp rot}
    &\dot{\bs{\gamma}}_t
    = 
    \bsA\bs{\gamma}_t + \bsB\bseta_\bs{\g}(t),\ 
    \bseta_\bs{\g}(t) \sim \mathcal{N}(\bzr, \bs{\S}_\bs{\g}).
\end{align}
By the model \eqref{eq: gp rot}, propagation and interpolation of the rotation state and its derivatives can be performed on flat Euclidean space $\R^3\times\R^3\times\R^3$ just like \eqref{eq: pva process}.
The kinematic models of the global states $(\rot_t, \angvel_t, \angacc_t)$ and local states $(\The_t, \Thed_t, \Thedd_t)$ are derived in Tab. \ref{tab: kinematic model derivation so3}, first by using the approximated method \cite{tang2019white, johnson2024continuous}, and then with our new closed-form approach.
In Remark \ref{sec: steam model}, we will further discuss the difference between approximated and closed-form models.

\subsection{GP with \texorpdfstring{\sepose}{SE3} Pose Representation} \label{sec: gp on se3}

The derivation of kinematic models for \sepose\ representation follows a similar procedure to the \srpose\ case in Tab. \ref{tab: kinematic model derivation so3}, therefore they are not presented side-by-side as in this Section.
Before detailing the derivation, let us recall some basic properties of \sepose. Let $\Xi = \ll(\The, \bsrho\rr)$ be the vector form of the Lie-algebra of $\SE$, where $\The \in \R^3$ is the rotation vector and $\bsrho \in \R^3$ is the translational part. The mappings $\Exp : \R^6 \to \SE$ and $\Log : \SE \to \R^6$ are defined as \cite{barfoot2024state}:
\begin{align}
    \Exp(\Xi)
    &\triangleq
    \begin{bmatrix}
        \Exp(\The) &\vbf{V}(\The)\bsrho\\
        \textbf{0}          &1
    \end{bmatrix}
    ,
    \label{eq: se3 exp defintion}
    \  
    \Log(\Exp(\Xi))
    =
    \begin{bmatrix}
        \The\\
        \bsrho
    \end{bmatrix}
\end{align}
where $\vbf{V}(\The) = \Jr(-\The)$ \cite{sola2018micro}.



Hence, between the time instances $t_k$ and $t_{k+1}$, we can convert between the local and global states in the same manner as Tab. \ref{sec: gptr on SO3}. Specifically given the global states $\tf$, $\twist$, $\wrench$ we can find the local state $\Xi$, $\Xid$, $\Xidd$ as
\begin{align}
    \Xi
    &\triangleq \Log(\tf_k^{-1}\tf),
    \label{eq: 0th derivative of xi}
    \\
    \Xid
    &\triangleq \SEJrInv(\Xi)\twist,
    \label{eq: 1st derivative of xi}
    \\
    \Xidd
    &=
    \frac{d\Xid}{dt}
    = \SEJrInv(\Xi)\frac{d\twist}{dt}
    +
    \frac{\partial }{\partial\Xi}\left[\SEJrInv(\Xi)\twist\right]
    \frac{d\Xi}{dt},
    \nonumber
    \\
    &\qquad\quad
    \triangleq
    \SEJrInv(\Xi)\wrench + \SEHp_1(\Xi, \twist)\Xid, 
    \label{eq: 2nd derivative of xi}
\end{align}
and from local states, we find the global states:
\begin{align}
    \tf
    &\triangleq \tf_k\Exp(\Xi),
    \\
    \twist
    &
    =
    \SEJr(\Xi)\Xid,
    \label{eq: velocity of se3}
    \\ 
    \wrench
    &= \frac{d\twist}{dt}
     = \SEJr(\Xi)\frac{d\Xid}{dt}
       +
       \frac{\partial }{\partial\Xi}\left[\SEJr(\Xi)\Xid\right]
       \frac{d\Xi}{dt}
    \nonumber
    \\
    &\qquad\quad \triangleq \SEJr(\Xi)\Xidd + \SEH_1(\Xi, \Xid)\Xid,
    \label{eq: acceleration of se3}
\end{align}
where $\twist = (\angvel, \bsnu)$ denotes the \textit{twist} and $\wrench = (\angacc, \bsbeta)$ is its derivative, $\SEJr(\Xi)$ is the right Jacobian of $\SE$, $\SEJrInv(\Xi)$ is its inverse $\SEJrInv(\Xi)$. The closed form of $\SEJr(\Xi)$, $\SEJrInv(\Xi)$, $\SEH_1(\Xi, \Xid)$ and $\SEHp_1(\Xi, \twist)$ are defined as follows:
\begin{align}
    &\SEJr(\Xi)
    \triangleq
    \begin{bmatrix}
        \Jr(\The) &\bzr\\
        \bsQ(\Xi) &\Jr(\The)\\
    \end{bmatrix},
    \\
    &
    \SEJrInv(\Xi)
    =
    \begin{bmatrix}
        \JrInv(\The)
        &\bzr\\
        \bsQp(\Xi) &\JrInv(\The)\\
    \end{bmatrix},
    \label{eq: Jrinv of SE3}
\end{align}
\begin{align}
    &\SEH_1(\Xi, \Xid)
    =
    \mypartder{\SEJr(\Xi)\Xid}{\Xi}
    \nonumber\\
    &=
    \renewcommand*{\arraystretch}{1.75}
    \begin{bmatrix}
        \mypartder{\Jr(\The)\Thed}{\The}
        &\bzr
        \\
        \mypartder{\bsQ(\Xi)\Thed}{\The} + \mypartder{\Jr(\The)\bsrhod}{\The}
        &\mypartder{\bsQ(\Xi)\Thed}{\bsrho}
    \end{bmatrix}
    \nonumber\\
    &=
    \begin{bmatrix}
        \SOH_1(\The, \Thed)
        &\bzr
        \\
        \SES_1(\Xi, \Thed) + \SOH_1(\The, \bsrhod)
        &\SES_2(\Xi, \Thed)
    \end{bmatrix}.
\end{align}

\begin{align}
    &\SEHp_1(\Xi, \twist)
    =
    \mypartder{\SEJrInv(\Xi)\twist}{\Xi}
    \nonumber\\
    &=
    \renewcommand*{\arraystretch}{1.75}
    \begin{bmatrix}
        \mypartder{\JrInv(\The)\angvel}{\The}
        &\bzr
        \\
        \mypartder{\bsQp(\Xi)\angvel}{\The} + \mypartder{\JrInv(\The)\bsnu}{\The}
        &\mypartder{\bsQp(\Xi)\angvel}{\bsrho}
    \end{bmatrix}
    \nonumber\\
    &=
    \begin{bmatrix}
        \SOHp_1(\The, \angvel)
        &\bzr
        \\
        \SESp_1(\Xi, \angvel) + \SOHp_1(\The, \bsnu)
        &\SESp_2(\Xi, \angvel)
    \end{bmatrix}, \label{eq: S'}
\end{align}
where $\bsQp(\Xi) \triangleq -\JrInv(\The)\bsQ(\Xi)\JrInv(\The)$,
and $\bsQ(\Xi)$ is detailed by Barfoot et al. in Eq. (8.91b) in \cite{barfoot2024state}. To keep this paper self-contained, we find an alternative form of $\bsQ(\Xi)$ as follows:
\begin{align} \label{eq: my Q definition}
    \bsQ(\Xi) = \bsQ(\The, \bsrho) = -\Exp(-\The)\SOH_1(-\The, \bsrho),
\end{align}
and $\SES_1$, $\SES_2$, $\SESp_1$, $\SESp_2$ are explained in Sec. \ref{sec: procedure}.

In contrast, Tang et al. \cite{tang2019white} previously used the Taylor expansion $\SEJrInv(\Xi) \approx \vbf{I} + \frac{1}{2}\Xi^\curlywedge $ to approximate \eqref{eq: 2nd derivative of xi} and \eqref{eq: acceleration of se3} as follows:
\begin{align}
    \Xidd &\approx \SEJrInv(\Xi)\wrench + \frac{1}{2}\Xid^\curlywedge\twist,
    \label{eq: xi approx} \\
    \dot{\twist} &\approx \SEJr(\Xi)\Xidd - \frac{1}{2}\SEJr(\Xi)\Xid^\curlywedge\twist.
    \label{eq: wrench approx}
\end{align}

\begin{remark} \label{sec: steam model}
    The approximations \eqref{eq: xi approx}, \eqref{eq: wrench approx} and are close to the closed form if $\Xi$ and $\Xid$ (or $\The$ and $\Thed$ for $\SO$) are small within the $\Dt$ interval, as remarked in \cite{tang2019white, johnson2024continuous}. This also informs us on when the two models would significantly diverge, i.e., in high-speed scenarios that entail larger relative motion in a MPI.
    In our framework, these approximations are also invoked when $\norm{\Xi}$ or $\norm{\The}$ is below a threshold to avoid numerical singularities. The approximated models can also be kept in effect at all times by a user-defined boolean parameter. When configured to use the approximated model with \sepose\ pose representation, the GPTR framework becomes equivalent to STEAM, except for some small implementation details (STEAM uses the left Jacobian convention, where ours uses right; GPTR uses the Ceres solver for GN optimization, while STEAM has a custom solver).
    In Sec. \ref{sec: experiment}, we shall conduct several experiments to demonstrate the benefit of the closed-form kinematic models over the approximated model based on \sepose, which stands in for the comparison with STEAM, as well as the approximated kinematic model for \srpose\ that was first formulated in \cite{johnson2024continuous}.
\end{remark}

\begin{remark}
    We notice in a very recent work \cite{barfoot2025integral}, Barfoot also derived a formula for derivatives in the kinematic models of general Lie groups based on some basic building blocks in the form of infinite series. In this paper, we take a different approach to derive the closed forms of \eqref{eq: local to global mappings}, \eqref{eq: glocal to local mappings}, \eqref{eq: 2nd derivative of xi}, \eqref{eq: acceleration of se3} by directly calculating the time derivatives of $\dot{\The}, \angvel$, $\dot{\Xi}$, $\twist$ from their already known closed forms, following the procedure in Sec. \ref{sec: procedure}. We also conduct experiments to establish the advantage of the closed-form model.
\end{remark}

\subsection{The Closed-form Derivative Procedure} \label{sec: procedure}

The calculation of $\SES_1$, $\SES_2$, $\SESp_1$, $\SESp_2$ in Sec. \ref{sec: gp on se3} as well as $\SOH_1$, $\SOHp_1$ in Sec. \ref{sec: gptr on SO3} all follow a similar procedure. We detail the calculation of $\SOH_1$ and $\SOHp_1$ in Appendix \ref{app: Jr JrInv H Hprime} as an illustration of the procedure.
However, due to the significantly larger number of computations involved in $\SES_1$, $\SES_2$, $\SESp_1$, $\SESp_2$, we forgo the explicit details and opt to delineate the procedure underlying these calculations.

Specifically, we notice that the closed forms of $\Exp(\The)$, $\Jr(\The)$, $\JrInv(\The)$, $\bsQ(\The, \bsrho)$ and $\bsQp(\The, \bsrho)$ all have a similar structure as follows:
\begin{equation} \label{eq: gf form}
    \bs{M} = \sum_{i=1}^{M} \f_i\ll(\The, \bsrho\rr) g_i\ll(\norm{\The}\rr),
\end{equation}
where $g_i(\cdot)$ consists of power and trigonometric terms of $\norm{\The}$ (see \eqref{eq: g10}-\eqref{eq: g32} for example), and $\f_i(\cdot, \cdot)$ are matrix polynomials of $\The^\wedge$ and $\bsrho^\wedge$ (see \eqref{eq: the f functions}). Given the closed forms of $g_i$ and $\f_i$, for a vector $\bsw$, calculating the closed-form Jacobian of $\bs{M}\bsw$ is equivalent to finding the sum of the Jacobians on each $f\bsw g$ term.
Let us take the calculation of $\SESp_1(\Xi, \angvel) = \mypartder{\bsQp(\The, \bsrho)\angvel}{\The}$ in \eqref{eq: S'} as an illustrative example:
\begin{align} \label{eq: derivative rule}
    \SESp_1
    &=
    \sum_{i=1}^{36}
    \Big[
        \f_{iu}(\bsu, \bsv, \bsw) g_i(u)\nonumber\\
        &\qquad\quad+ \f_i(\bsu, \bsv)\bsw g_i^{(1)}(u)\bar{\bsu}^\top
    \Big]
    \Big|_{\bsu = \The, \bsv = \bsrho, \bsw = \angvel},
\end{align}
where $u \triangleq \norm{\bsu}$ and $\bar{\bsu} \triangleq \bsu/u$, $g^{(n)}_i(u) \triangleq \frac{\partial^n g_i(u)}{\partial u^n}$ and $\f_{iu}(\bsu, \bsv, \bsw) \triangleq \mypartder{\f_{i}(\bsu, \bsv)\bsw}{\bsu}$.

Due to the amount of $\f g$ terms involved in $\bsQ(\Xi)$ and $\bsQp(\Xi)$ and the page constraint, we shall not detail the closed forms of $g_i^{(n)}$, $\f_{iu}$  as with the \srpose\ case in Appendix \ref{app: Jr JrInv H Hprime}. We refer the readers to our source code for these details.



\section{Unified Gaussian Process Trajectory Representation} \label{sec: trajectory representation}

{
As explained earlier, when switching $\tf$ between and $(\rot, \pos)$, one has to re-implement all the factors in each ME task based on the selected representation. Obviously, this process is far from desirable, especially if one seeks to benchmark between different representations. Hence, in this section, we propose a trajectory representation scheme that allows users to build the same factors for the ME tasks that are invariant to the underlying kinematics.
}

\subsection{Trajectory Representation} 
A continuous-time trajectory represented by GP can be defined by three sets of user-defined parameters:
\begin{itemize}
    \item The motion prior interval $\Dt \in \R^+$.
    \item A set of support states $\{\hat{\X}_k\}_{k=0}^{K}$, where $\hat{\X}_k \triangleq (\hat{\rot}_k, \hat{\angvel}_k, \hat{\angacc}_k, \hat{\pos}_k, \hat{\vel}_k, \hat{\acc}_k) \in \SO\times\R^3\times\R^3\times\R^3\times\R^3\times\R^3$. Each support state represents the state estimates of the robot at the time instance $t_k = t_0 + k\Dt$.
    \item The covariance matrices $\bs{\S}_\bs{\nu}, \bs{\S}_\bs{\g} \in \R^{3\times3}$ as defined in \eqref{eq: pva process} and \eqref{eq: gp rot}.
\end{itemize}

{If \srpose\ GP is chosen, then
we assume that the \textit{rotational states} $\rot_k, \angvel_k, \angacc_k$ evolve by the model \eqref{eq: gp rot}, and the translational states $ \pos_k, \vel_k, \acc_k$ follow \eqref{eq: pva process}.
On the other hand, if the $\SE$ GP is chosen, we can reshuffle the states internally to form the global states $\tf_k, \twist_k, \wrench_k$ and assume that the local states $\Xi, \Xid, \Xidd$ evolve by \eqref{eq: pva process}. This encapsulation strategy is visualized in Fig. \ref{fig: encapsulation}.
Based on the choice of the kinematic model, the interpolation and \textit{intrinsic Jacobians} are calculated accordingly without affecting the ME factors.

\begin{figure}
    \centering

    \begin{subfigure}[b]{\linewidth}
        \includegraphics[width=\linewidth]{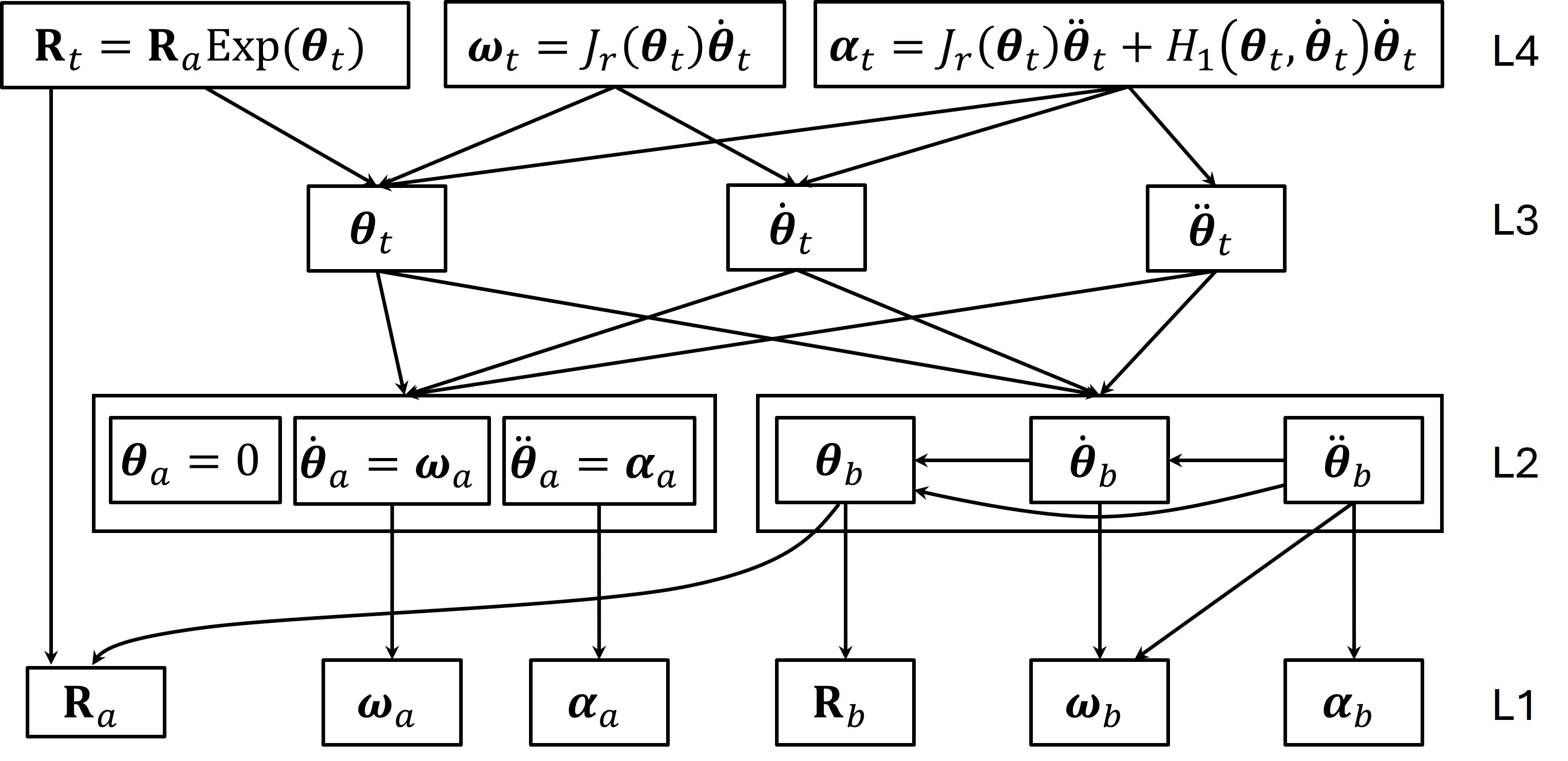}
        \caption{The hierarchy of manifold variables in $\SO$ GP. We group the variables into four levels, and the goal is to find the Jacobian of variables in level L4 w.r.t to variables in level L1. Every arrow from variable $v_i$ to $v_j$ represents a non-zero partial derivative $\Jcb{v_i}{v_j}$. To avoid clustering, every arrow that goes into the sub-group $\gamma_a$ or $\gamma_b$ represents three actual arrows, each pointing to each variable in the subgroup. The intrinsic Jacobians describing these dependencies are detailed in Appendix \ref{sec: instrinsic jacobians srpose}.}
        \vspace{0.5cm}
        \label{fig: jacobian heirarchy}
    \end{subfigure}
    \begin{subfigure}[b]{\linewidth}
        \includegraphics[width=\linewidth]{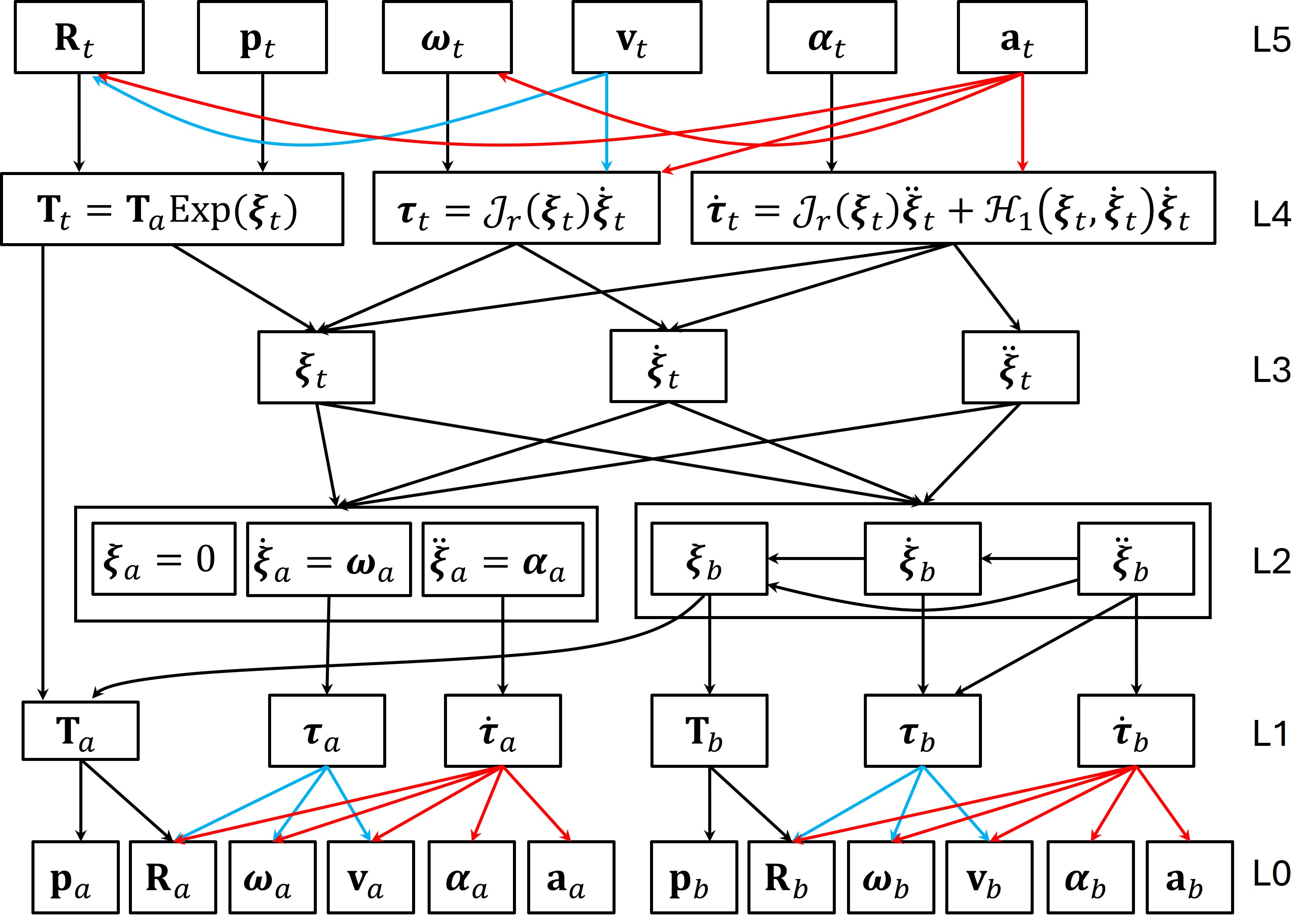}
        \caption{The hierarchy of manifold variables in the \sepose\ GP. We use color to help distinguish some variables with many dependencies. The structure is the same from L4 to L1, only that we now have new variables in L5 and L0. The Jacobians between L5 and L4 are detailed in \eqref{eq: R wrt T}--\eqref{eq: a wrt wrench}, and the Jacobians between L1 and L0 are detailed in \eqref{eq: T twist wrench wrt to rospva}--\eqref{eq: L1 L0 Jcb} of Appendix \ref{sec: instrinsic jacobians sepose}. The use of these L5-L4 and L1-L0 Jacobians is the key to the unified trajectory representation in this work.}
        \label{fig: jacobian heirarchy se3}
    \end{subfigure}
    
    \caption{ The dependency graphs describing the hierarchy of manifold variables for $\SO$ GP (Fig. \ref{fig: jacobian heirarchy}) and $\SE$ GP.}
    \label{fig: encapsulation}
\end{figure}


}

{
\begin{remark}
    Our choice of $(\hat{\rot}_k, \hat{\angvel}_k, \hat{\angacc}_k, \hat{\pos}_k, \hat{\vel}_k, \hat{\acc}_k)$ as the basic state representation comes from the experience that $\rot$, $\angvel$, $\pos$, $\vel$ can often be directly measured by sensors, and thus more intuitive, whereas ${\tf}_k, {\twist}_k$ appear to be some recombinations of ${\rot}_k, {\angvel}_k, {\pos}_k, {\vel}_k$. In fact, we observe that the $\rot$ state is the source of all complexities regarding state estimation.
    In theory, we can also use $(\hat{\tf}_k, \hat{\twist}_k, \hat{\wrench}_k)$ as the basic state representation and $(\hat{\rot}_k, \hat{\angvel}_k, \hat{\angacc}_k, \hat{\pos}_k, \hat{\vel}_k, \hat{\acc}_k)$ as the internally shuffled states.
\end{remark}
}


\subsection{Regression}

Given an observation or sensor measurement $\Z$ that is coupled with the state $\X_t$ via some observation model $h(\Z, \X_t, \dots, \eta) = 0$, where $\eta$ represents the noise or uncertainty.
The regression problem can be formalized as:
\begin{equation} \label{eq: least square}
    \argmin_{\{\hat{\X}_m\}_{m=0}^{K}} \sum_{\Z} \norm{r(\Z_t, \hat{\X}_t, \dots)}_\bsW^2,
\end{equation}
where $\norm{r}^2_\bsW \triangleq r^\top \bsW r$, and $r \triangleq h(\Z, \hat{\X}_t, \dots, \eta)\vert_{\eta=0}$ is the residual and $\bsW$ is a weight based on the inverse of the covariance of $\eta$ (also known as the information matrix). 
Crucial to the solving of \eqref{eq: least square}, one needs to calculate the Jacobian of the residual $r(\Z, \hat{\X}_t,\dots)$ over the states estimate $\hat{\X}_a$, $\hat{\X}_b$ where $\left[t_a,\ t_b\right] \ni t$. By using the chain rule, we have
\begin{equation}
    \frac{\partial r}{\partial \hat{\X}_a} = \frac{\partial r}{\partial \hat{\X}_t} \frac{\partial \hat{\X}_t}{\partial \hat{\X}_a},\ \frac{\partial r}{\partial \hat{\X}_b} = \frac{\partial r}{\partial \hat{\X}_t} \frac{\partial \hat{\X}_t}{\partial \hat{\X}_b}.
\end{equation}
The Jacobian $\frac{\partial r}{\partial \hat{\X}_t}$ depends on each specific residual.
However, $\frac{\partial \hat{\X}_t}{\partial \hat{\X}_a}$ and $\frac{\partial \hat{\X}_t}{\partial \hat{\X}_b}$, which we call the \textit{intrinsic Jacobians}, follow specific formulas {based on the underlying pose representation ($\SO \times \R^3$ or $\SE$)}, as shown in Fig. \ref{fig: encapsulation}.
Calculating the closed form of these Jacobians can be quite involved.
However, they can substantially reduce computational cost compared to the automatic differentiation approach, sometimes up to 80\% in some of our tests.
We provide a systematic procedure for these calculations in Appendix \ref{sec: instrinsic jacobians}, believing that the process can inspire a deeper understanding of Lie algebra. We have also verified our derivations by using the automatic derivative feature in Ceres, which uses the jet number representation for finding the derivative of a function\footnote{\url{http://ceres-solver.org/automatic_derivatives.html}}. 

\color{black}

\subsection{Motion prior}

Between every two consecutive support states $\hat{\X}_a$ and $\hat{\X}_b$ on the trajectory (we denote $a = k$ and $b = k+1$ for simpler notations), we add a motion prior factor that is specialized for the underlying pose representation. This factor is crucial for the convergence of the CTME  problem based on GP, as it constrains the feasible trajectories to those conforming to the dynamic models \eqref{eq: pva process} and \eqref{eq: gp rot}, effectively acting as a smoothing factor.

For \srpose, we define the residual $\res_\M$ and the weight $\bseta_\M$ as follows:
\begin{align*} 
    \res_\M
    &=
    \begin{bmatrix*}[l]
        \res_\g
        \\        
        \res_\nu
    \end{bmatrix*}
    \triangleq
    \begin{bmatrix*}[l]
        \hat{\bs{\g}}_b - \bsF(\Dt)\hat{\bs{\g}}_a
        \\        
        \hat{\bs{\nu}}_b - \bsF(\Dt)\hat{\bs{\nu}}_a
    \end{bmatrix*},
    \\
    \bsW_\M
    &= \begin{bmatrix*}[l]
        \bsQ_\bs{\g}(\Dt) &\bzr
        \\        
        \bzr &\bsQ_\bs{\nu}(\Dt)
    \end{bmatrix*}^{-1}.
\end{align*}

Similarly, for the $\SE$ pose representation, we can define the residual as
\begin{align*} 
    \res_\M
    \triangleq \hat{\bs{\Xi}}_b - \ll[\bar{\bsF}(\Dt) \otimes \vbf{I}_6\rr]\hat{\bs{\Xi}}_a
    ,\ 
    \bsW_\M 
    = \bsQ_\bs{\Xi}(\Dt)^{-1}.
\end{align*}

As the motion prior is specialized for the chosen pose representation, we also provide a function overload to help the user quickly construct the factors without having to calculate the Jacobians for each type of motion prior above.

\section{GP-based CTME Prototypes} \label{sec: scenarios}

In this section, we will present some GP-based CTME scenarios. Sections \ref{sec: VIO}, \ref{sec: UWB IMU}, \ref{sec: mlcme} will present the general formulation of these CTME schemes for visual, ranging, and LiDAR scenarios.

\subsection{Visual-Inertial Motion Estimation} \label{sec: VIO}

In this first example, we demonstrate the application of GP-based CTME in calibrating visual-inertial sensor suites. Let us define the state estimates as follows:
\begin{align}
    \hat{\T}
    &\triangleq
    \left(\dots, \hat{\X}_{m}, \dots, \hat{\bias}_a, \hat{\bias}_\o, \hat{\grav}, {}^\im_{\ca_0}\hat{\tf},{}^\im_{\ca_1}\hat{\tf}\right),\label{eq: vistates}
    \\
    \hat{\X}_{m} &= (\hat{\rot}_{m}, \hat{\angvel}_{m}, \hat{\angacc}_{m}, \hat{\pos}_{m}, \hat{\vel}_{m}, \hat{\acc}_{m}),\label{eq: vicp}
\end{align}
where $\hat{\bias}_a$ and $\hat{\bias}_\o$ are accelerometer and gyroscope biases, respectively. $\hat{\X}_m$ are the support states of the trajectory estimate of the IMU sensor. $\hat{\grav}$ denotes the estimated gravity vector. $\{{}^\im_{\ca_i}\hat{\tf}\}_{i=0}^1$ are the transformation from camera $i$ to IMU frame, i.e., the extrinsics. Hence, the motion estimation and calibration using known landmarks in a batch-wise manner is formulated as the minimization of the following cost function:
\begin{align} \label{eq: cost function vie}
    f(\hat{\T})
    = &\sum_{(\hat{\X}_k, \hat{\X}_{k+1})}\norm{\res_\M}^2_{\vbf{W}_\M}
    \nonumber\\
    &\qquad\qquad+
    \sum_{\mathcal{C}_t}\norm{\res_\mathcal{C}}^2_{\vbf{W}_\mathcal{C}}
    +    \sum_{\I_t}\norm{\res_\I}^2_{\vbf{W}_\I},
\end{align}
where $\res_\M$, $\res_\mathcal{C}$, and $\res_\mathcal{I}$ are the \textit{motion prior}, projection, and IMU residuals, respectively. $\vbf{W}_\M$, $\vbf{W}_\mathcal{C}$, and $\vbf{W}_\I$ are the corresponding information matrices (inverse of covariance). $\\SEC_t$ and $\I_t$ refer to the visual and inertial observations obtained during the optimization period.

\subsection{Ranging-based Motion Estimation} \label{sec: UWB IMU}

Next, we demonstrate batch optimization of range measurements with GP-based CTME. Specifically, we aim to estimate the following states:
\begin{align}
    \hat{\T}
    &\triangleq
    \left(\hat{\X}_{1}, \dots \hat{\X}_{m}\right),
\end{align}
with support state $\hat{\X}_{m}$ defined similarly to \eqref{eq: vistates} and \eqref{eq: vicp}. The problem formulation for this scenario becomes minimizing the following cost function:
\begin{align} \label{eq: cost function UWB}
    f(\hat{\T})
    = \sum_{(\hat{\X}_k, \hat{\X}_{k+1})} \norm{\res_\M}^2_{\vbf{W}_\M}
    +
    \sum_{\U_t}\norm{\res_\U}^2_{\vbf{W}_\U}
\end{align}
where $\res_\M$, $\res_\U$ denote the motion prior and range factors, respectively; $\vbf{W}_\M$, $\vbf{W}_\U$ are the corresponding information matrices. $\U_t$ denotes a range observation within the sliding window.

\color{black}

\subsection{Multi-LiDAR Coupled-Motion Estimation} \label{sec: mlcme}

\begin{figure}
    \centering
    \footnotesize
    \begin{overpic}[width=0.75\linewidth,
                   ]{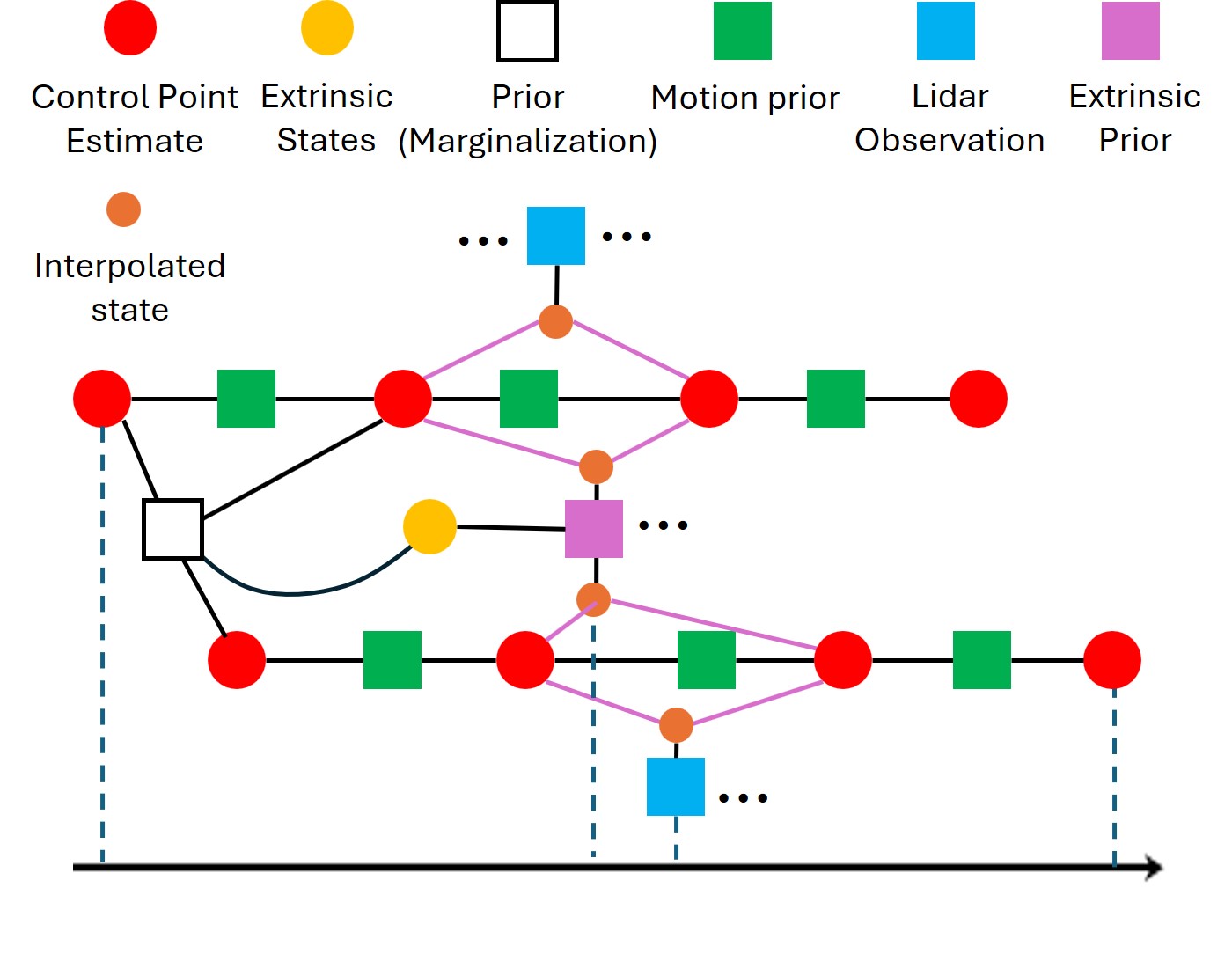}
                   \put(08, 02){ $t_s$}
                   \put(90, 02){ $t_e$}
                   \put(52, 02){ $t$}
                   \put(30, 52){ ${}_0\hat{\X}_{k}$}
                   \put(37, 18){ ${}_1\hat{\X}_{k}$}
                   \put(27, 39){ ${}^{\frL_0}_{\frL_1}\hat{\tf}$}
    \end{overpic}
\caption{The factor graph of the GP-based multi-LiDAR coupled-motion estimation scheme. We have two trajectories corresponding to two LiDARs.
Each square represents one factor, and the lines indicate the coupled states. The extrinsic prior factors reinforce the \textit{a priori}.
}
\label{fig: factor graph}
\end{figure}

In the previous scenarios, only one trajectory was represented by the GP.
Inspired by \cite{pervsic2021spatiotemporal}, we also demonstrate a GP-based multi-trajectory optimization scheme. Different from \cite{pervsic2021spatiotemporal}, which performed batch optimization for sensors with overlapping field of view (FOV).
In this paper, we shall investigate a Maximum A Posteriori (MAP) estimation scheme of two LiDARs that do not have any overlapping FOV.
Moreover, while motion distortion is often considered undesirable in discrete-time motion estimation schemes, in this example, we would also like to demonstrate that CTME with GP representation can directly extract the motion in the LiDAR distortion \textit{without relying on IMU} \cite{zheng2024traj, zheng2025traj}.
Specifically, we define the following state estimates on a sliding window:
\begin{align}
    \hat{\T}
    &\triangleq
    \left(\dots {}_0\hat{\X}_{m},
          \dots
          {}_1\hat{\X}_{m}, 
          \dots,
          {}^{\frL_0}_{\frL_1}\hat{\tf}
          \right),
    \\
    {}_i\hat{\X}_{m} &= ({}_i\hat{\rot}_{m}, {}_i\hat{\angvel}_{m}, {}_i\hat{\angacc}_{m}, {}_i\hat{\pos}_{m}, {}_i\hat{\vel}_{m}, {}_i\hat{\acc}_{m}),\label{eq: cpx}
\end{align}
where ${}_i\hat{\X}_{m}$ are the support states on the sliding window of trajectory $i$ and ${}^{\frL_0}_{\frL_i}\hat{\tf} \triangleq ({}^0_i\rot, {}^0_i\pos) \in \SE$ are the transformations between LiDAR $0$ and LiDAR $i$. The state estimates can be optimized via the cost function \eqref{eq: cost function}:
\begin{align} \label{eq: cost function}
    &f(\hat{\T})
    = \norm{\res_\P}^2_{\bsW_\P}
    +
    \sum_{({}_i\hat{\X}_k, {}_i\hat{\X}_{k+1})} 
    \norm{\res_\M}^2_{\bsW_\M}
    \nonumber\\
    &\quad+
    \sum_{\L_t}\norm{\res_\L}^2_{\bsW_\L}
    +
    \sum_{\tau}\norm{\res_\E}^2_{\bsW_\E},
\end{align}
where $\res_\P$, $\res_\M$ are respectively the prior from marginalization and the motion prior, $\res_\L$ is the residual of LiDAR-based point-to-plane factors, and $\res_\E$ is the residual of \textit{extrinsic prior factor}. $\bsW_\P$, $\bsW_\M$, $\bsW_\L$, $\bsW_\E$ are correspondingly their information matrices. Fig. \ref{fig: factor graph} illustrates the factor graph based on this cost function with two LiDARs.

\section{Experimental Validation} \label{sec: experiment}

\subsection{Visual-Inertial Motion Estimation}

\begin{figure}
    \centering
    \adjustbox{trim={0.15\width} {0.3\height} {0.15\width} {.3\height},clip}{\includegraphics[width=0.6\textwidth]{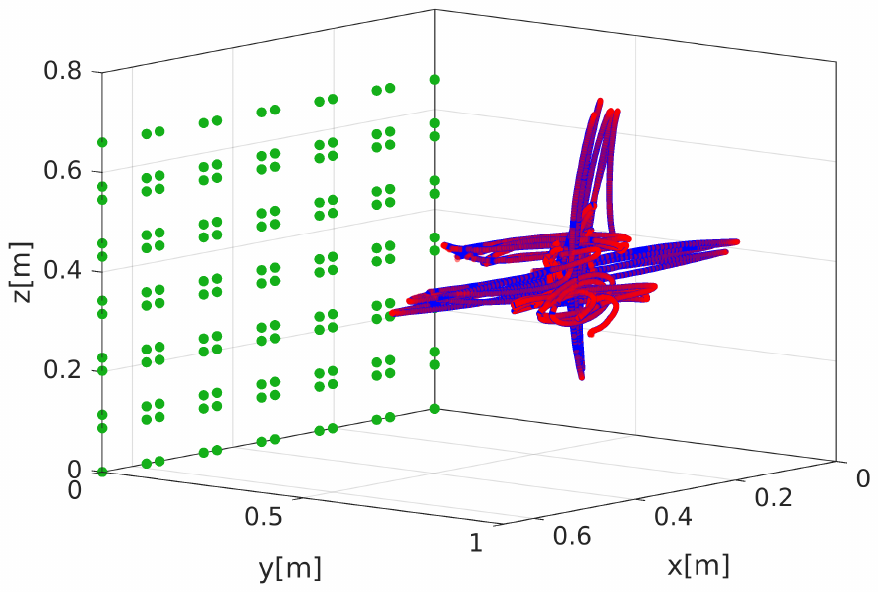}}
\caption{Illustration of the estimated trajectory (blue) and ground truth (red) in the visual-inertial calibration given known landmark positions depicted as green dots.}
\label{fig: vitraj}
\end{figure}

We first investigate the proposed CTME scheme in a VI optimization scheme based on a calibration problem using the real-world dataset in~\cite{sommer2020efficient}, which was originally provided in~\cite{schubert2018vidataset}. A stereo-visual-inertial sensor suite is assembled and captures images at $20$ Hz and inertial data at $200$ Hz. A checkerboard is set up w.r.t.\ the world frame, and the ground-truth trajectory during calibration was recorded at about $100$ Hz using a motion capture system.
We use the double sphere camera model \cite{usenko2018double} in this experiment. Fig. \ref{fig: vitraj} visualizes the visual landmarks, ground truth, and one trajectory estimate in our experiments.

Though the data is captured at one velocity profile, we modify it to emulate motion at various higher velocity profiles. Specifically, suppose $\breve{\angvel}_t$ and $\breve{\angacc}_t$ are the IMU angular velocity and acceleration, with $t$ as the timestamp of the data recorded. By scaling the time stamp and the IMU data by a factor $\lambda$, i.e.
\begin{align}
    t' = t/\lambda,\ \breve{\angvel}_{t'} = \lambda\breve{\angvel}_t,\ \breve{\angacc}_{t'}=\lambda^2\breve{\angacc}_{t},
\end{align}
the new data with timestamp $t'$ is equivalent to being captured at a velocity that is $\lambda$-time faster.

\begin{figure*}
    \centering
    \includegraphics[width=0.9\linewidth]{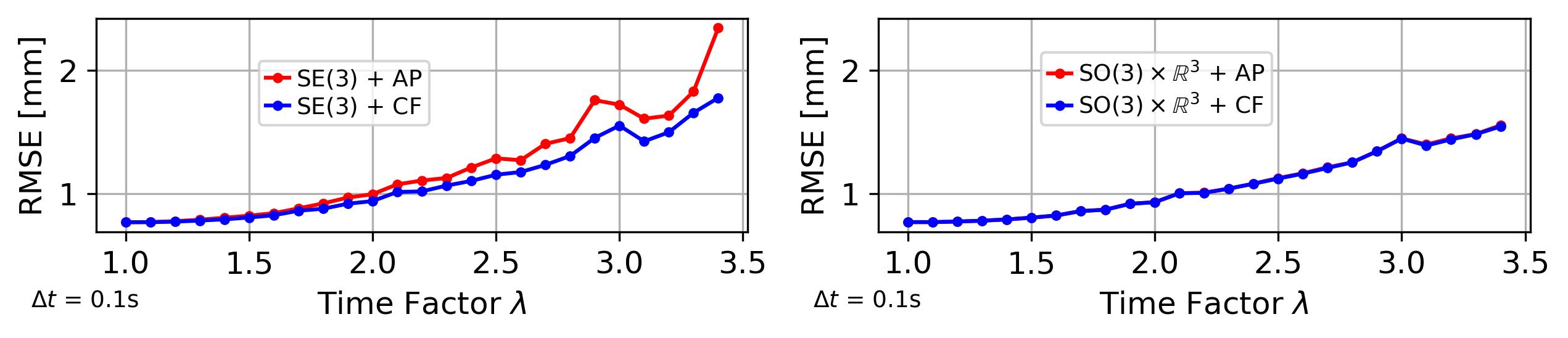}
    \includegraphics[width=0.9\linewidth]{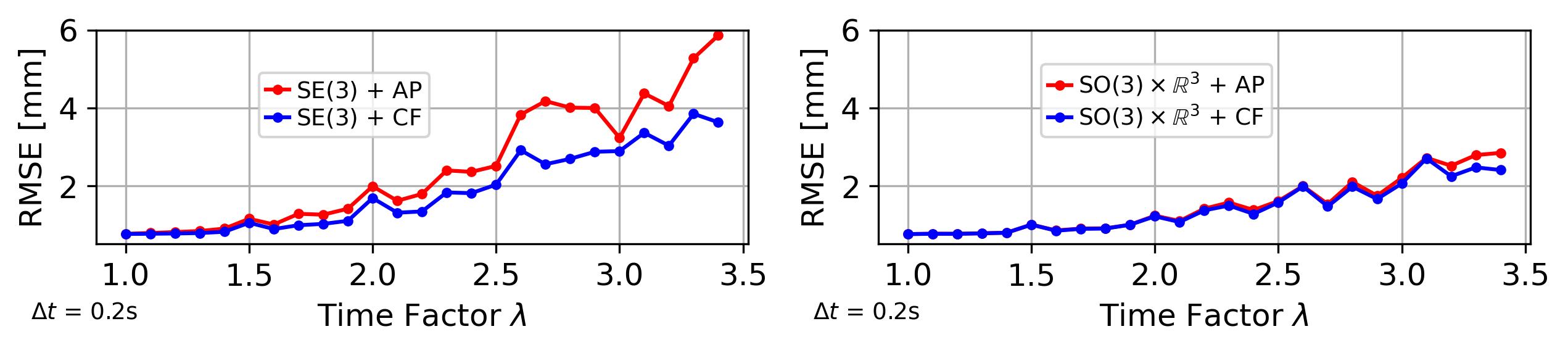}
    \caption{ The RMSE of VI experiments with 0.1s (top) and 0.2s (bottoms) MPIs under various time factors. In the legends, \srpose\ and \sepose\ refer to the chosen pose representations of trajectory estimate, while AP and CF respectively refer to our implementation by GPTR with the approximated kinematics model and the closed-form model.
    A STEAM-based implementation would be equivalent to the \sepose+AP settings. We can see that at low speed, the CF and AP models have similar accuracy. However, at higher speed or sparser $\Dt$, the AP model starts gaining higher error compared to the CF model. This divergence of performance between CF and AP can be viewed more easily for the \sepose\ (top left and bottom left) than the \srpose.}
    \label{fig: so3xr3 vi rmse}
\end{figure*}

The RMSEs of different models across the time scales are presented in Fig. \ref{fig: so3xr3 vi rmse}.
We can see that approximated (AP) and closed-form (CF) models perform similarly at low speed. However, at higher speed, the AP model starts picking up more error. This effect can be viewed more clearly for the \sepose\ pose representation.

\begin{figure*}
    \centering
    \includegraphics[width=0.9\linewidth]{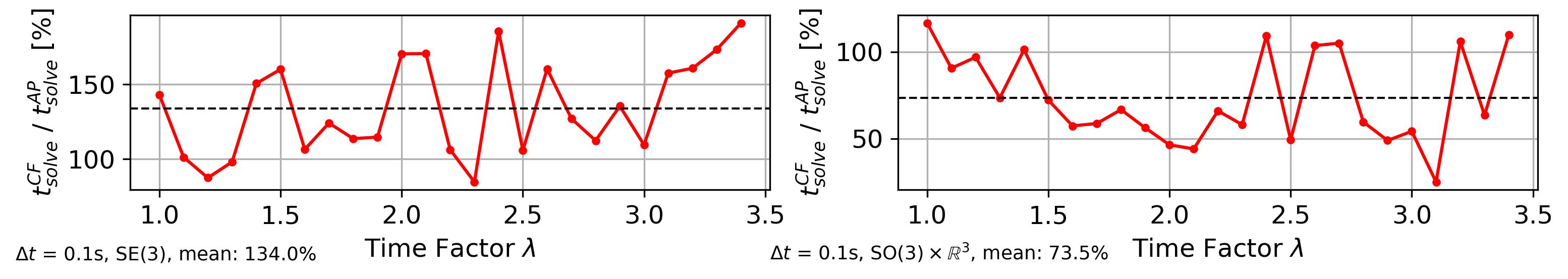}
    \includegraphics[width=0.9\linewidth]{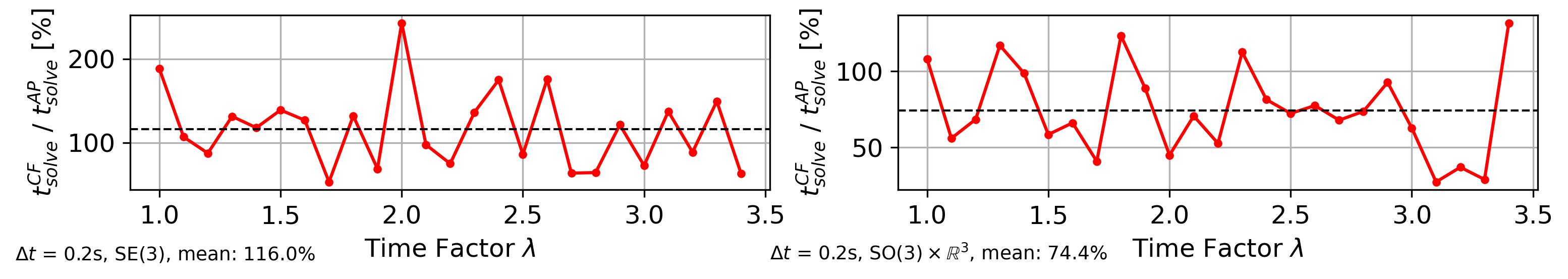}
    \caption{The ratio of solve time between the CF and AP models corresponding to the VI experiments in Fig. \ref{fig: so3xr3 vi rmse}.
    Notice that there are several experiments where the CF model has much lower solve time.
    This is because the Ceres solver finds that the problem has converged after fewer iterations.}
    \label{fig: so3xr3 vi solve time}
\end{figure*}

The ratio of solve time between CF and AP models are presented Fig. \ref{fig: so3xr3 vi solve time}. Interestingly, we note there are several experiments where the solve time of CF model is lower than the AP model for the \srpose\ representation, leading to a lower mean. We found that this is because the Ceres optimizer converges with fewer iterations when using the CF kinematic model.

\textbf{Key finding for VI Motion Estimation:} In summary, for VI optimization, the CF model has similar accuracy at low speed and outperforms AP at higher velocities. It also shows that the \srpose+CF model can require fewer iterations to converge, thus improving the efficiency.

\subsection{Ranging-based Motion Estimation} \label{sec: uwb experiment}

To further examine the importance the pose representation for the ME task, we conduct a range-based experiment with two types of ground truth motions as follows.

\subsubsection{Split Trajectory}
\begin{figure}
    \centering
    \begin{subfigure}[b]{0.49\linewidth}
        \includegraphics[width=\linewidth]{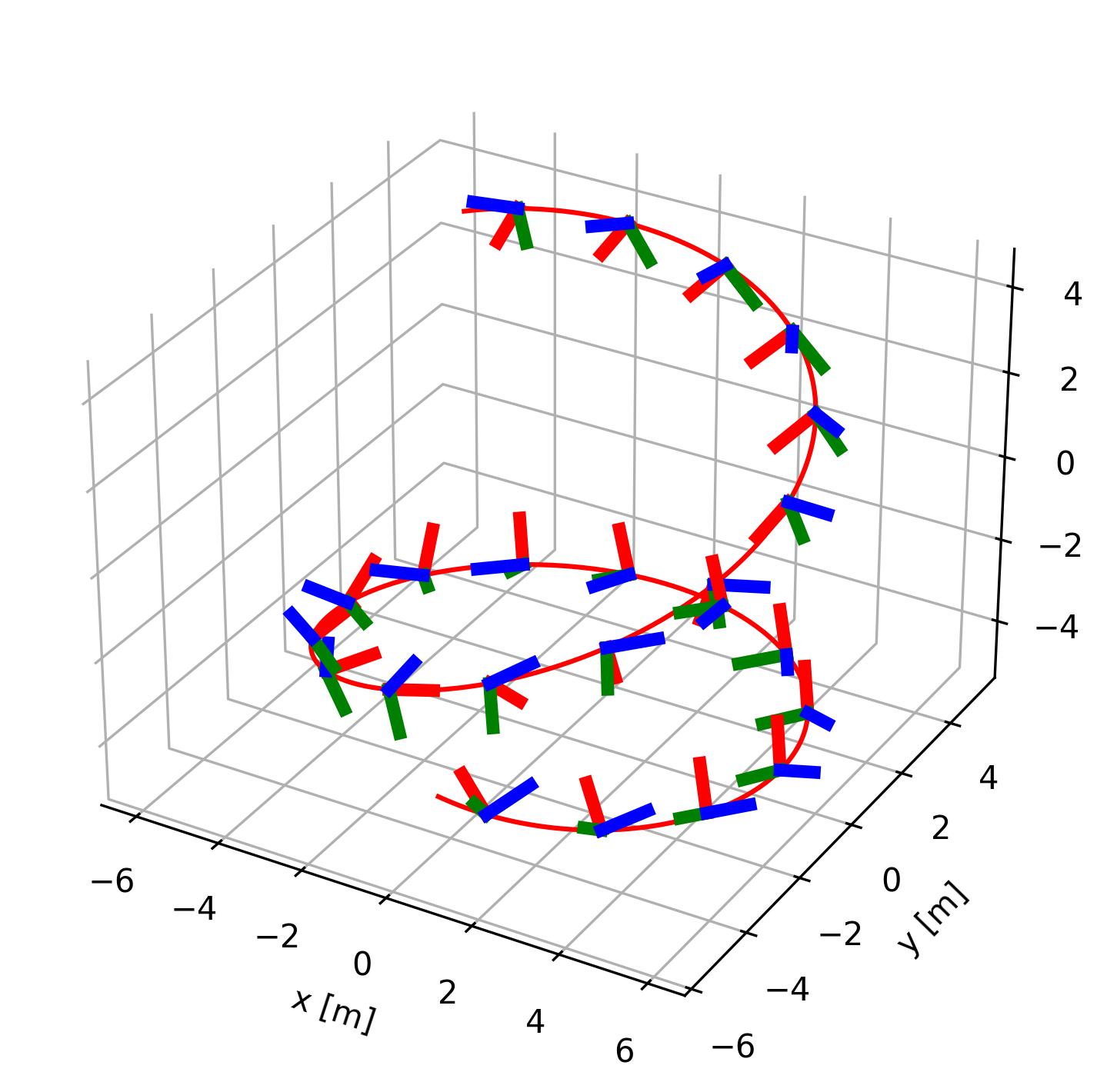}
        \caption{Split trajectory}
        \label{fig: split trajectory gtr}
    \end{subfigure}
    \begin{subfigure}[b]{0.49\linewidth}
        \includegraphics[width=\linewidth]{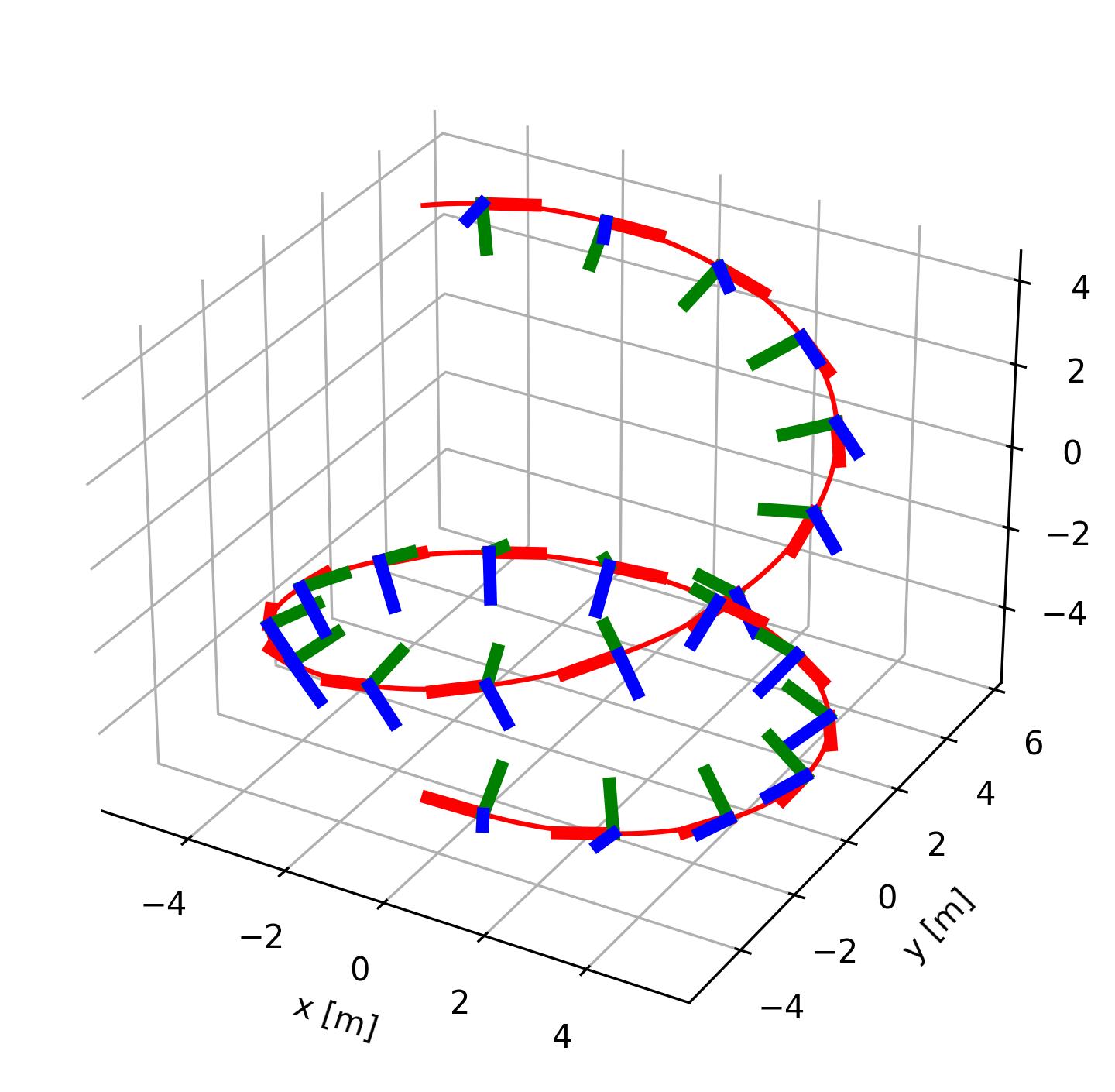}
        \caption{Non-split trajectory}
        \label{fig: non-split trajectory gtr}
    \end{subfigure}
    \caption{Ground truth trajectories for experiments with UWB-only experiments in Sec. \ref{sec: uwb experiment}.}
\end{figure}

We generate a so-called \textit{split trajectory} as ground truth as follows:
\begin{align} \label{eq: so3xr3 gt traj}
    \rot_t &= \Exp\ll(\bs{\theta}_t\rr),\ 
    \bs{\theta}_t = \begin{bmatrix}
        \pi/2\cos(\Omega t + 57)\\
        \pi/2\sin(\Omega t + 57)\\
        \pi\sqrt{3}/2\sin(\Omega t / 3 + 43)
    \end{bmatrix}
    ,
    \nonumber\\
    \pos_t &= \begin{bmatrix}
        5\sin(0.45 t + 43)\\
        5\cos(0.45 t + 43)\\
        5\cos(0.15 t + 57)
        \end{bmatrix},
\end{align}
where $\Omega$ is varied in different experiments. This trajectory is illustrated in Fig. \ref{fig: split trajectory gtr}.
\begin{figure*}
    \centering

    \begin{subfigure}[b]{0.9\linewidth}
        \includegraphics[width=\linewidth]{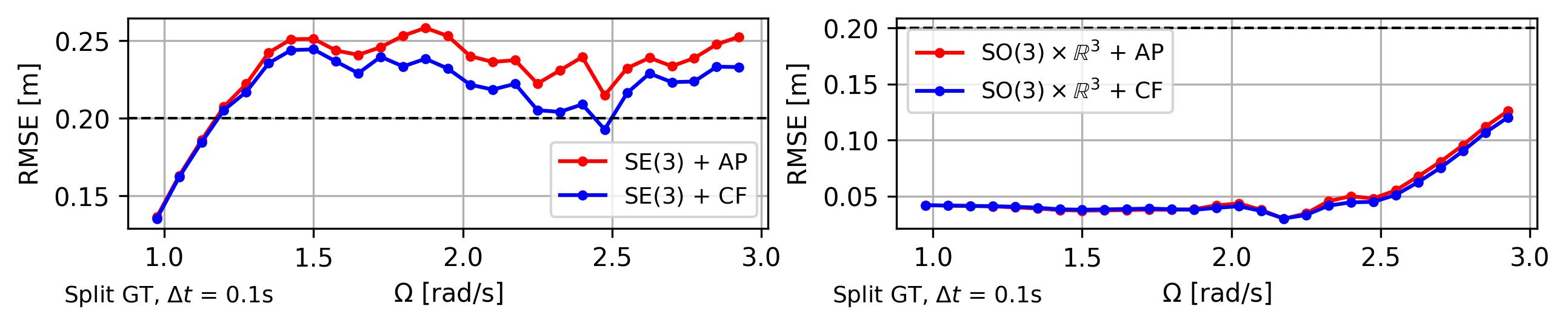}
        \caption{Experiments with split groundtruth trajectory \eqref{eq: so3xr3 gt traj}}
        \label{fig: uwb so3xr3 traj rmse}
    \end{subfigure}
    
    \begin{subfigure}[b]{0.9\linewidth}
        \includegraphics[width=\linewidth]{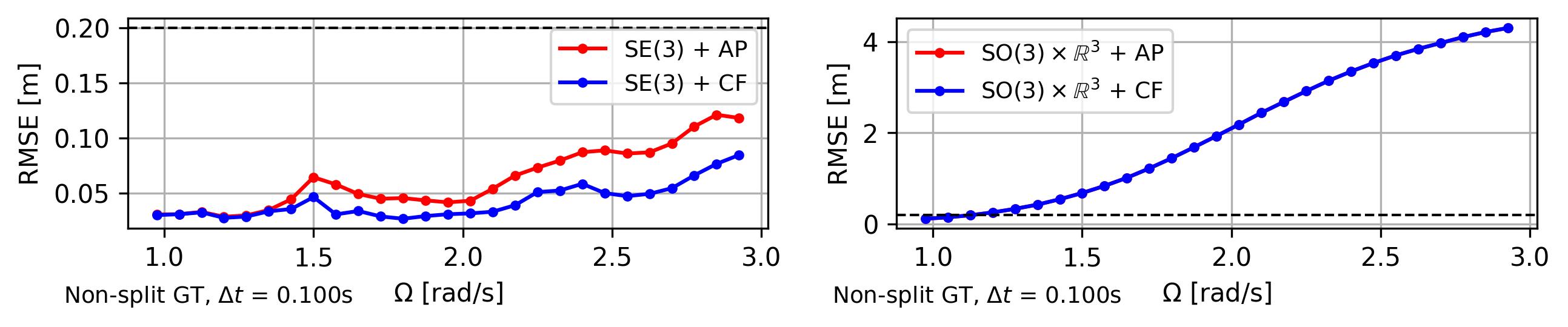}
        \caption{Experiments with non-split groundtruth trajectory \eqref{eq: se3 gt traj}}
        \label{fig: uwb se3 traj rmse}
    \end{subfigure}
    
    \caption{ RMSE of the range-only optimization experiments with different values of frequency parameter $\Omega$. In the legends, \srpose\ and \sepose\ refer to the pose representation used, while AP and CF refer to the approximated and closed-form kinematic models. For the split ground truth trajectory in \eqref{eq: so3xr3 gt traj}, the \sepose+X models (left plots) start having an RMSE above 0.2m quickly with larger $\Omega$, while the \srpose+X representation (right plots) maintains an RMSE below 0.2m. On the contrary, in Fig. \ref{fig: uwb se3 traj rmse}, the \srpose\ pose representation (left plots) starts losing track of the true trajectory quickly when $\Omega$ increases, while the \sepose\ representation maintains smaller RMSE in all experiments. This demonstrates the importance of choosing the appropriate pose representation.}

    \label{fig: uwb exp rmse}
    
\end{figure*}


\begin{figure*}
    \centering
    \includegraphics[width=0.9\linewidth]{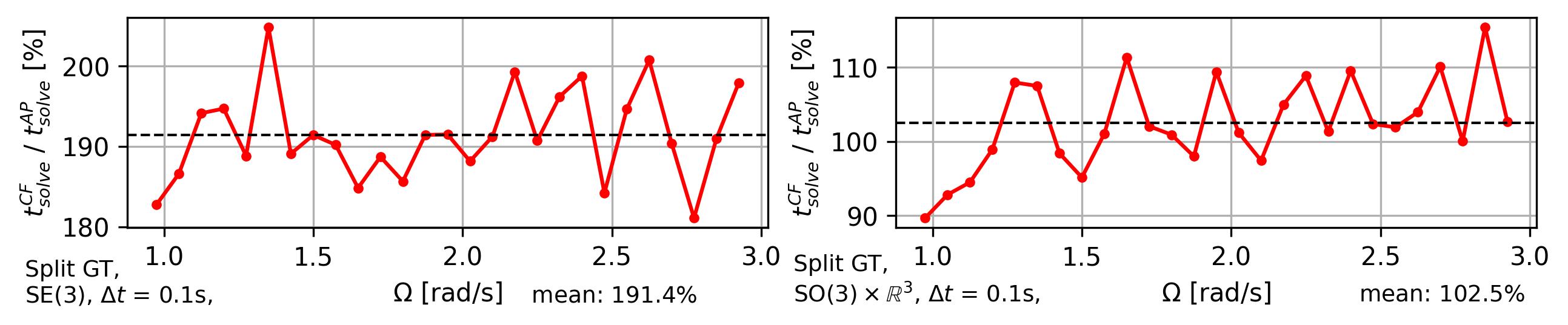}
    \caption{The ratio of solving time by closed-form kinematics model and the AP model in the ranging-based optimization problem, using \srpose\ (left) and \sepose\ (right) representations.}
    \label{fig: so3xr3 uwb solve time}
\end{figure*}

We assume that there are two UWB tags that are offset from the body center by ${}^\frB\vbf{x} \in \{(-0.2, 0, 0), (0.2, 0, 0)\}$. At 50ms interval, each UWB tag ranges to four anchors whose coordinates are $( 10.0,  10.0, 0.5)$, $(-10.0,  10.0, 2.5)$, $(-10.0, -10.0, 0.5)$, $( 10.0, -10.0, 2.5)$. The measurement is corrupted by a zero-mean gaussian noise with 0.05-$m^2$ variance.

We vary the frequency $\Omega$ of the trajectory ground truth and the MPI $\Delta t$ of the trajectory estimate throughout the experiments. For each variation, we initialize a 20s-long trajectory estimate with the true pose corrupted by Gaussian noise with a 0.2-$radian^2$ variance for rotation state and $0.5-m^2$ variance for position state. All 1st and 2nd derivatives are initialized as zeros. We then optimize the cost function with UWB and motion prior factors by the ceres solver with the maximum number of iterations set at 50. The RMSE of all experiments are shown in Fig. \ref{fig: uwb so3xr3 traj rmse}, which are discussed below.

From Fig. \ref{fig: uwb so3xr3 traj rmse}, we find that at $\Delta t = 0.1s$, the AP model starts picking up more error than CF at higher $\Omega$. We notice that the difference between AP and CF are less ample in \srpose+X setting than the \sepose+X setting.
Notice that the RMSEs of the \sepose+X models also exceed 0.2m very quickly, while \srpose+X models can keep the error below 0.2m well as $\Omega$ increases.
These results are evidence for the benefit of choosing the proper pose representation, as well as the closed-form model over the approximation model.


\subsubsection{Non-split Trajectory}
To confirm the benefits of choosing the appropriate pose representation with regards to the type of motion, we generate another \textit{non-split trajectory} as follows:
\begin{align} \label{eq: se3 gt traj}
    \pos_t &= \begin{bmatrix}
        5\sin(\Omega t + 43)\\
        5\cos(\Omega t + 43)\\
        5\cos(\Omega t / 3 + 57)
        \end{bmatrix},
        \ \rot_t = [\textbf{e}_x, \textbf{e}_y, \textbf{e}_z],
        \nonumber
        \\
        \textbf{e}_x &= {\dot{\pos}_t/\norm{\dot{\pos}_t}},
        \ 
        \textbf{e}_z = \ll({\pos_t/\norm{\pos_t}}\rr)^\wedge\textbf{e}_x,
        \ 
        \textbf{e}_y = \textbf{e}_z^\wedge\textbf{e}_x.
\end{align}
This trajectory is visualized in Fig. \ref{fig: non-split trajectory gtr}. It has the same translational component as Fig. \ref{fig: split trajectory gtr}, however the x-axis of the body frame is always tangent to the trajectory. This trajectory resembles that of a fixed-wing aircraft, or a wheeled vehicle. All of the parameters are kept the same as in the experiments with the split trajectory, except that we flip the choice of the trajectory representation from \srpose\ to \sepose. The results of these experiments are reported in Fig. \ref{fig: uwb se3 traj rmse}.

As can be seen in Fig. \ref{fig: uwb se3 traj rmse}, on this new trajectory, the \sepose\ representation clearly has an advantage over \srpose, as the \srpose\ RMSE quickly increases when $\Omega$ increases. Similar to the previous case, we also find that the AP model starts getting higher error than the CF model at larger $\Omega$.
This again demonstrates the benefit of the CF kinematic model for trajectories with higher velocities. Moreover, though it is not obvious, this also demonstrates the flexibility of our framework as all experiments are carried out in one single program with the same UWB factor class, and the only difference is the settings on the trajectory object.


\begin{figure*}
    \centering

    \begin{subfigure}[b]{0.8\linewidth}
        \includegraphics[width=\linewidth]{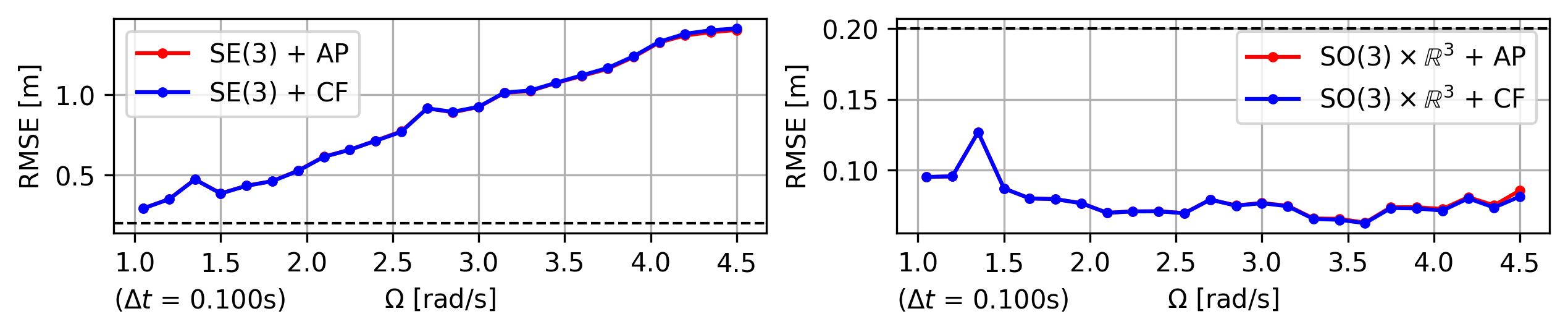}
        \includegraphics[width=\linewidth]{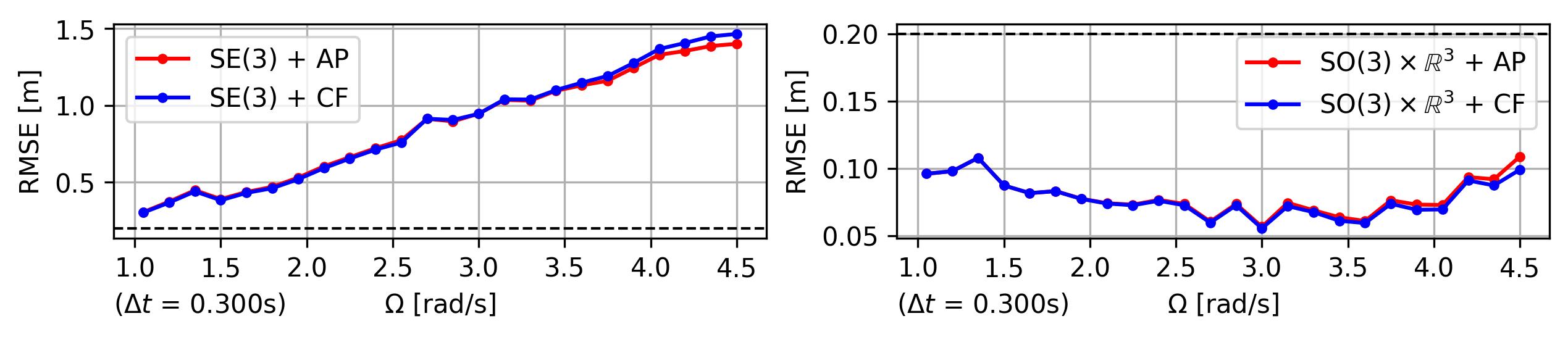}
        \caption{ RMSE of the lidar-based optimization at different MPIs under the split trajectory defined in \eqref{eq: so3xr3 gt traj}.}
        \label{fig: lidar so3xr3 traj rmse}
    \end{subfigure}
    
    \begin{subfigure}[b]{0.8\linewidth}
        \includegraphics[width=\linewidth]{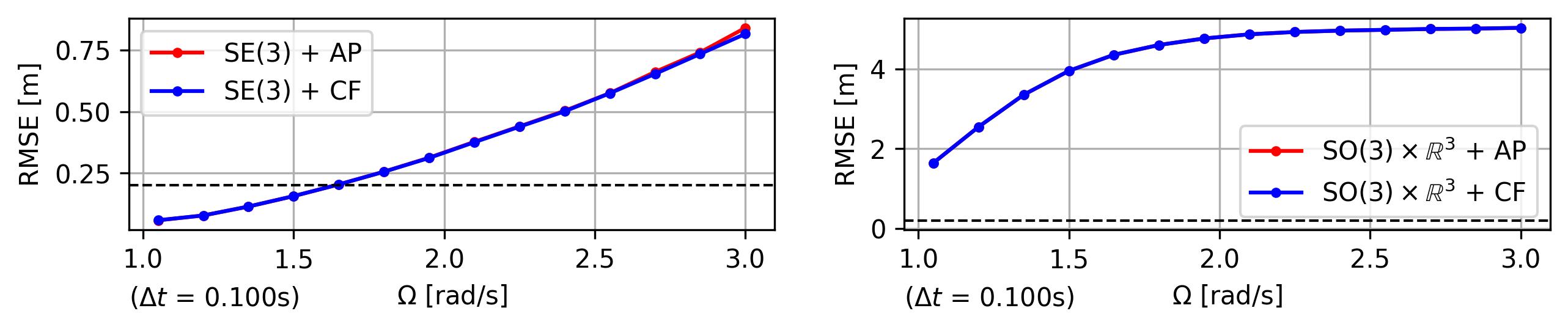}
        \includegraphics[width=\linewidth]{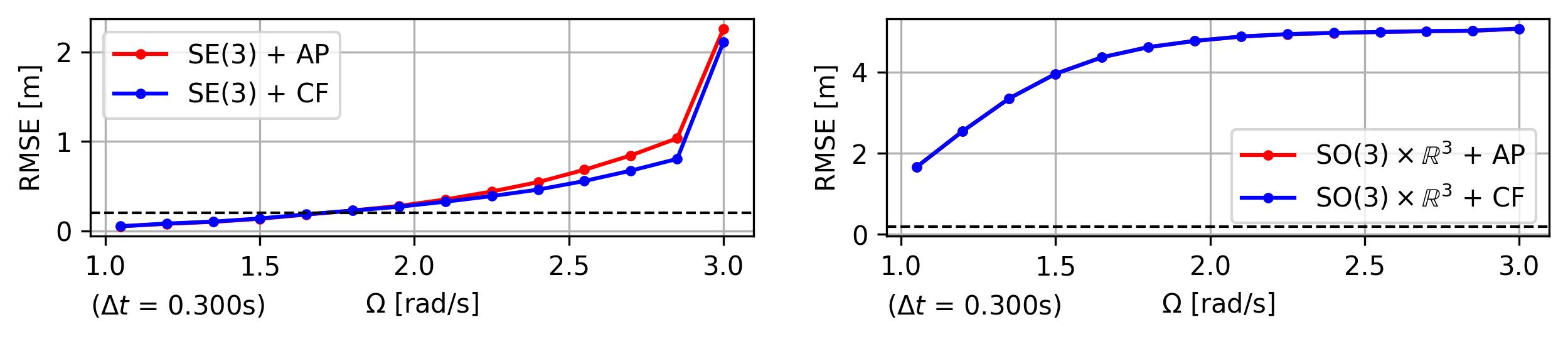}
        \caption{ RMSE of the lidar-based optimization at different MPIs under the non-split trajectory defined in \eqref{eq: se3 gt traj}.}
        \label{fig: lidar se3 traj rmse}
    \end{subfigure}
    
    \caption{ RMSE of the lidar-based batch optimization experiments.}
    \label{fig: lidar exp rmse}  
\end{figure*}


Fig. \ref{fig: so3xr3 uwb solve time} presents the ratio of the solving time for the experiments on the trajectory \eqref{eq: so3xr3 gt traj} using different methods. We can see that on average, the CF models incurs a small increase in computation against the AP model for the \srpose\ pose representation, while the \sepose\ representation incurs \~90\% more computation time. 

\textbf{Key findings for Range-Based Motion Estimation:} In summary, the CF kinematic model consistently outperforms AP under higher motion profiles, especially when the pose representation aligns with the trajectory type.
These results highlight the effectiveness and flexibility of the proposed framework in ranging-based motion estimation.

\subsection{Lidar-based CTME}

\subsubsection{Batch optimization}

We first conduct several experiments similar to Sec. \ref{sec: uwb experiment}, where the UWB factors are replaced with point-to-plane factors. Instead of ranging to specific anchors, we simulate the lidar observations by finding the intersection of several lidar rays with the surrounding walls of a $12m\times12m\times8m$ chamber. The rate of ray tracing is 300Hz. This optimization can be found at the core of bundle-adjustment schemes for lidar-based mapping \cite{liu2021balm, li2024pss}. The results are shown in Fig. \ref{fig: lidar so3xr3 traj rmse} and Fig. \ref{fig: lidar se3 traj rmse}.

From Fig. \ref{fig: lidar so3xr3 traj rmse}, we find that the CF and AP models still stay close, even when we have extended to higher values of $\Omega$. When increasing a MPI to $0.3s$, we can start noticing a difference in \srpose+X models with the split ground truth. For the split experiment, the \sepose+X models still diverge quickly. Conversely, for the non-split ground truth, the \srpose+X models diverge quickly, while \sepose+X models can maintain estimation capacity for longer.
Based on this experiment, we conclude that the CF may not have a clear advantage when applying with point-to-plane factors. It remains consistent with previous experiments that there is a benefit in choosing the appropriate pose representation for different motion types.

\textbf{Key finding for Lidar-based CTME:} In this experiment, the LiDAR-based CTME experiments show no clear performance difference between the closed-form and AP models.

In the next parts we will focus more on practical applications of GP-based CTME and compare its performance against existing methods of lower orders.

\color{black}

\subsubsection{Real world handheld setup}

\begin{figure}
    \centering
    \includegraphics[width=\linewidth]{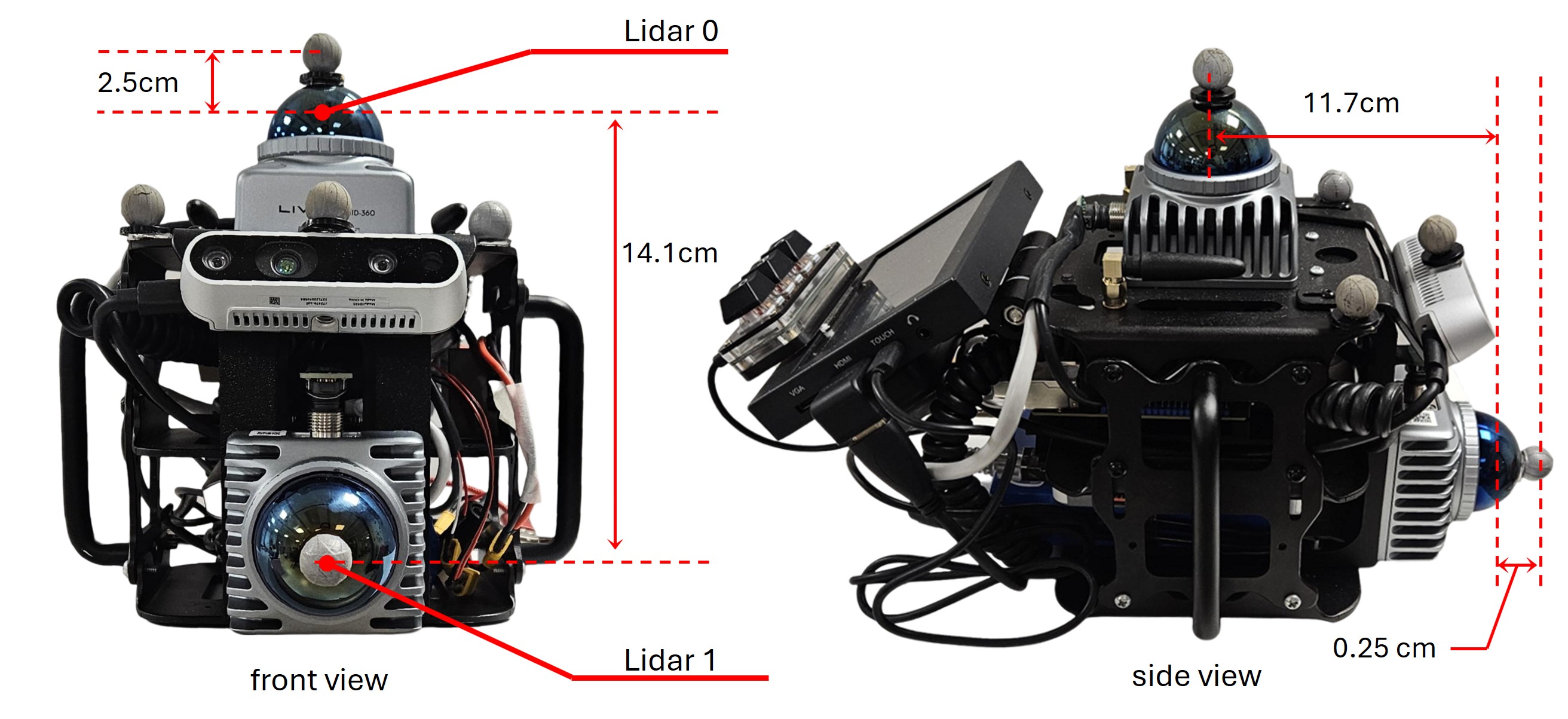}
\caption{The two-lidar sensor suite. Based on the illustration in the Mid360 Datasheet, we assume the origin of the lidar's coordinate system coincides with the center of the half-spheres.}
\label{fig: cathhs setup}
\end{figure}

In this section we conduct an experiment of the MLCME scheme on a handheld setup with two lidars.
{Three state-of-the-art lidar-only lidar-odometry (LO) methods are selected for comparison, namely I2EKF-LO \cite{yu2024i2ekf},  Traj-LO \cite{zheng2024traj}, and {CTE-MLO}\cite{shen2024cte}. In the frontend, all methods use a similar point-to-plane factor with slightly different feature association strategies: I2EKF-LO, Traj-LO, and GPTR use KNN search, while CTE-MLO uses a voxel map.
In the backend, we disable the mapping process and only use a fixed prior map for all methods. The similarities in both frontend and backend allow us to focus the comparison on the trajectory representation and the kinematic model. I2EKF uses a second-order WNOJ; Traj-LO is a constant velocity model with \sepose\ pose representation; CTE-MLO uses a second-order $\SO$ GP and a third-order GP in the translational part.}
We first test the multi-lidar calibration scheme on a real-world setup as shown in Fig. \ref{fig: cathhs setup}, which consists of two Livox-Mid360 lidars. We use the datasheet illustration, as well as the hardware CAD model, to deduce the ground truth for the extrinsics. {The ground truth extrinsics are directly used to enable multi-lidar capability of CTE-MLO}. The trajectory ground truth is obtained from a motion capture system, with careful compensation for the reflective markers and the lidar's frame of reference.

\begin{figure}
    \centering
    \includegraphics[width=\linewidth]{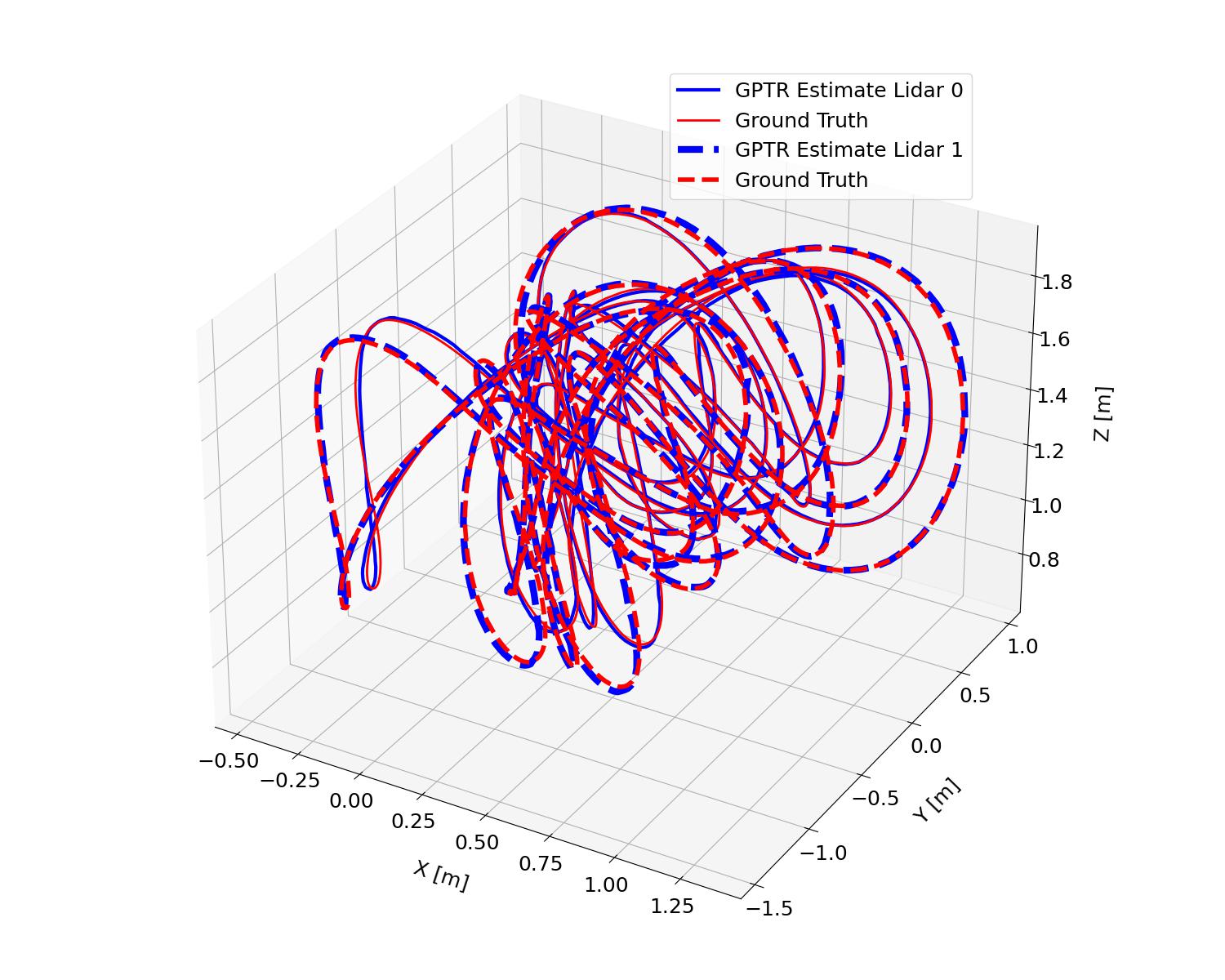}
\caption{Trajectory estimate for each lidar by the MLCME scheme vs the ground truth.}
\label{fig: cathhs 3d plot}
\end{figure}

\begin{figure*}
    \centering
    \includegraphics[width=\linewidth]{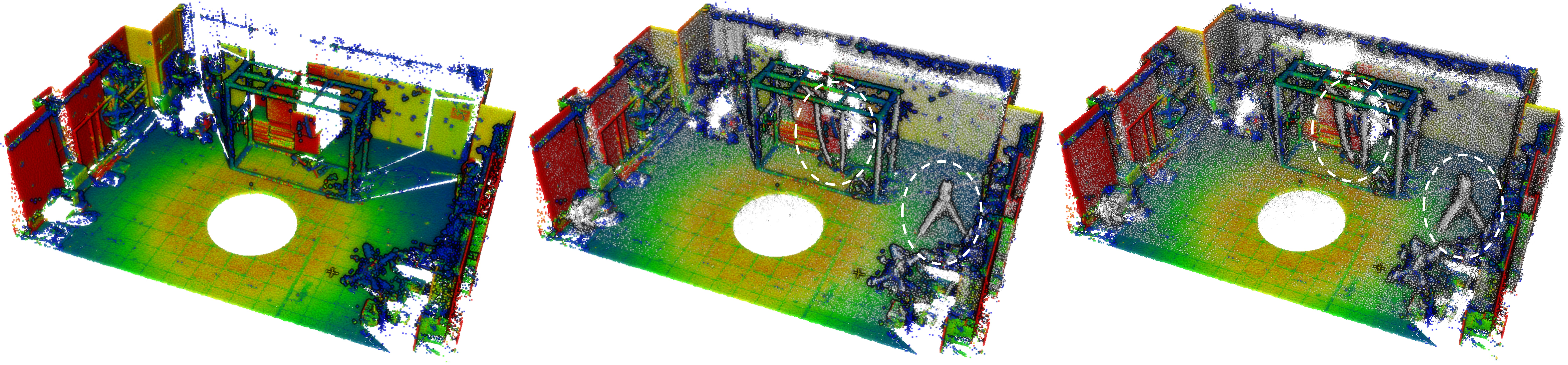}
    \caption{Left: the prior map. Middle: the lidar point clouds, undistorted and overlaid on the prior map. Right: the distorted raw point cloud over the prior map.}
    \label{fig: cathhs pointcloud registration}
\end{figure*}

\begin{table}
    
    \centering
    \renewcommand{\arraystretch}{1.5}
    \begin{threeparttable}
    \caption{ Comparing the RMSE of trajectory estimate by the MLCME scheme in GPTR and other methods with the handheld setup data.} \label{tab: cathhs mlcme ape}
    \begin{tabular}{ccccccc}
        \hline\hline
        \mr{2}{*}{\textbf{Seq\#}}
        &\mr{2}{*}{\bf{Lidar}}
        &\mr{2}{*}{\bf{I$^2$EKF}}
        &\mr{2}{*}{\bf{Traj}}
        &\mr{2}{*}{\bf{CTEM}}
        &\mc{2}{c}{\bf{GPTR}}
        \\\cline{6-7}
        & & & & & {\scriptsize \sepose} & {\scriptsize \srpose} \\\cline{6-7}
        \hline
        \mr[c]{2}{*}{0} & 0 & 0.0476 & 0.027 & 0.043 & \ul{0.0236} & \bf{0.0233} \\
         & 1 & 0.6793 & 0.0442 & 0.0417 & \ul{0.0285} & \bf{0.0272} \\
        \cline{1-7}
        \mr[c]{2}{*}{1} & 0 & 0.0647 & 0.0278 & 0.042 & \ul{0.0234} & \bf{0.0233} \\
         & 1 & 0.043 & 0.1319 & 0.0418 & \ul{0.0316} & \bf{0.031} \\
        \cline{1-7}
        \mr[c]{2}{*}{2} & 0 & 0.0477 & 0.0422 & 0.0554 & \ul{0.0401} & \bf{0.0394} \\
         & 1 & x & 0.0568 & 0.0567 & \ul{0.0452} & \bf{0.0445} \\
        \hline\hline
    \end{tabular}
    \begin{tablenotes}
        \small
        \item Note: All values are in [m] unit. \textbf{Traj} and \textbf{CTEM} are shorthand for Traj-LO \cite{zheng2024traj} and CTE-MLO \cite{shen2024cte}. The methods in GPTR use the closed-form kinematic model.
    \end{tablenotes}
    \end{threeparttable}
\end{table}

\begin{figure}
    \centering
    \includegraphics[width=\linewidth]{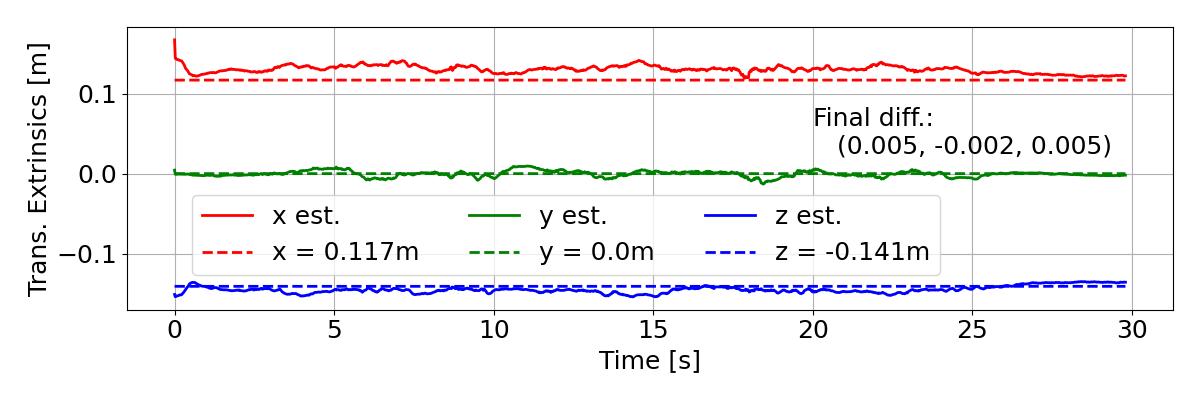}
    \includegraphics[width=\linewidth]{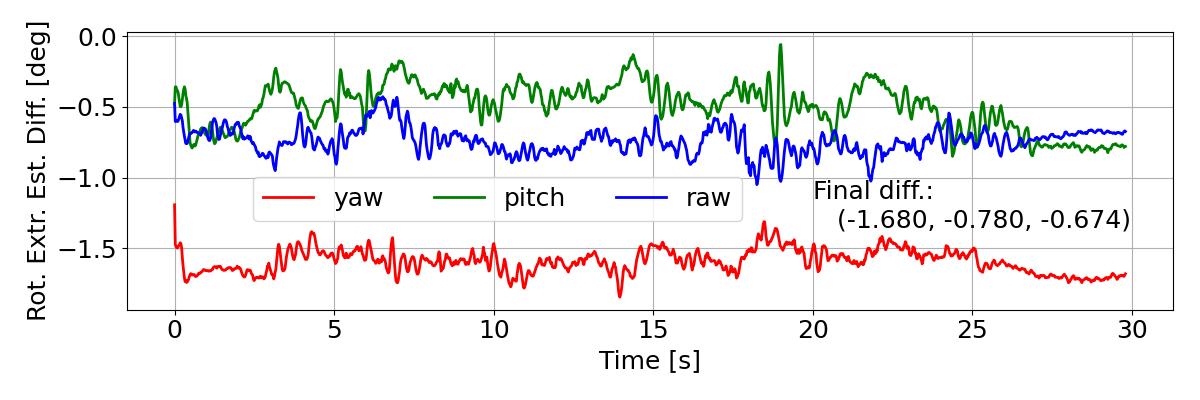}
\caption{The extrinsic estimate result between LiDAR 0 and LiDAR 1.}
\label{fig: cathhs extrinsic estimation}
\end{figure}

Figure \ref{fig: cathhs 3d plot} shows the trajectory estimates of the lidars and the ground truth. It can be seen that in this experiment, the motion is quite aggressive, where the setup is swung around in the experiment area. We calculate the APE of the trajectory estimates from I2EKF, Traj-LO, CTE-MLO and our method, and report it in Tab. \ref{tab: cathhs mlcme ape}. {The table shows that the MLCME scheme using the \srpose\ model receives the highest accuracy, followed by the \sepose.
We also notice CTEMLO's accuracy is close to GPTR's thanks to its use of multiple lidars, which enjoys more robustness thanks to extended field of view. In contrast, I$^2$EKF and Traj-LO, being single-lidar methods, have higher errors on each lidar's trajectory.} In Fig. \ref{fig: cathhs extrinsic estimation}, we can also see that the extrinsic parameters converge to some constant values which are within 1.5cm and 2$^o$ of our assumed ground truth. In Fig. \ref{fig: cathhs pointcloud registration}, we also show the pointclouds undistorted by the CT trajectory overlaid on top of the prior map, compared to that without distortion.

\subsubsection{Synthetic data}
\begin{figure}
    \centering
    \begin{overpic}[width=\linewidth,
                   ]{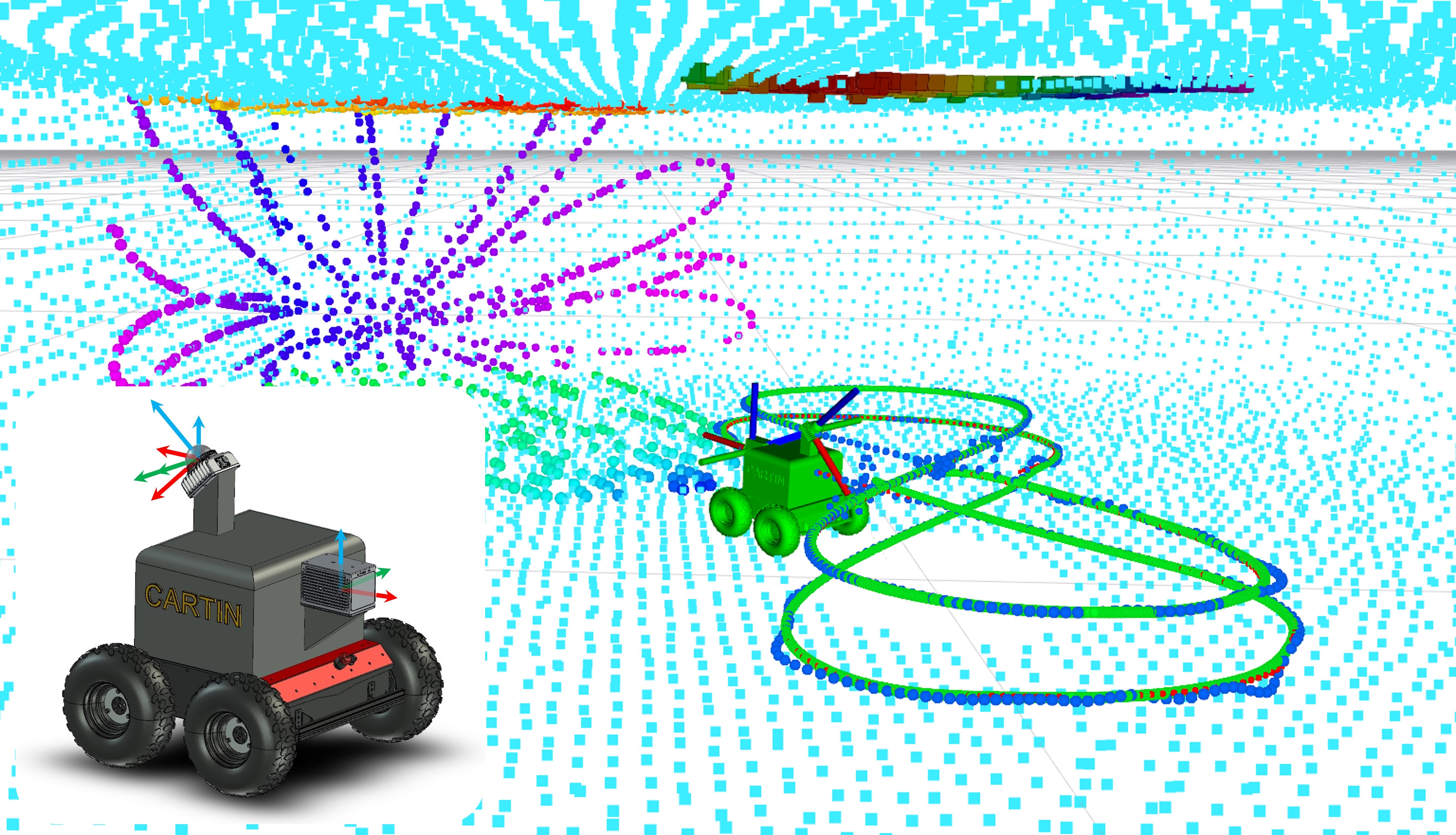}
                   \put(05, 22){ \footnotesize $\frL_0$}
                   \put(07, 27){ \footnotesize $\frB$}
                   \put(27, 17){ \footnotesize $\frL_1$}
    \end{overpic}
\caption{Simulated setup with two LiDARs mounted on top of a ground robot. The ground robot moves in a room with six flat sides. The two LiDARs have no overlap. The GPTR trajectory and I2EKF-LO estimates are marked by green and blue, respectively, while the ground truth trajectories are in red.}

\label{fig: simulation setup}
\end{figure}

In generating the synthetic dataset, we assume two LiDARs, 0 and 1, are mounted on a vehicle (see Fig. \ref{fig: simulation setup}) and their extrinsics are set as a user-defined parameter. The trajectory of the vehicle is predetermined by a continuous function where {$\pos_t = (2\sin(\Omega t), 2\sin(\Omega t)\cos(\Omega t), 0.75)$ and $\rot_t$ is the same as in \eqref{eq: se3 gt traj}. The parameter $\Omega$ is varied in different sequences}. We then sample the vehicle's pose every 2e-5s, convert it to the LiDAR's pose, perform ray-tracing to the walls of a 6m $\times$ 6m room to obtain the distance, add a Gaussian noise of {}{5cm} variance to the range, and finally calculate the 3D point.
The coordinate of LiDAR 0 and LiDAR 1 in the robot body frame expressed in a yaw-pitch-roll-x-y-z tuple is set as:
\begin{align*}
    &{}^\frB_{\frL_0}\tf = (45^\circ, 0, 0, 0, 0, 0)
    \\
    &{}^\frB_{\frL_1}\tf
    =
    \begin{cases}
        (180^\circ, 0, 0, -0.5, 0, -0.25),\ \text{if\ } t \notin (10, 20)\\
        (180^\circ, 0, 0, -0.5, 0, -0.35),\ \text{otherwise}.
    \end{cases}
\end{align*}
Hence, the two LiDARs' FOV do not overlap. From ${}^\frB_{\frL_0}\tf$ and ${}^\frB_{\frL_1}\tf$ we can calculate the transform between $\frL_0$ and $\frL_1$ as
\begin{align*}
    {}^{\frL_0}_{\frL_1}\tf
    &= (180^\circ, 45^\circ, 0, \frac{-1}{4\sqrt{2})}, 0, \frac{-3}{4\sqrt{2}})
    \\
    &\approx (180^\circ, 45^\circ, 0, -0.177, -0.530),\ \text{if\ } t \notin (10, 20)
\end{align*}
and
\begin{align*}
    {}^{\frL_0}_{\frL_1}\tf
    &= (180^\circ, 45^\circ, 0, \frac{-3}{20\sqrt{2})}, 0, \frac{-17}{20\sqrt{2}})
    \\
    &\approx (180^\circ, 45^\circ, 0, -0.106, -0.601),\ \text{if\ } t \in (10, 20),
\end{align*}
which is the ground truth for the MLCME scheme.
The change of extrinsic mimics a scenario where the sensor mechanical holder may slip during operation.
The LiDAR initial poses are set with 10cm translation and $5^\circ$ rotation errors. The MPI is chosen to be {0.04357s}.

\begin{figure}
    \centering
    \includegraphics[width=\linewidth]{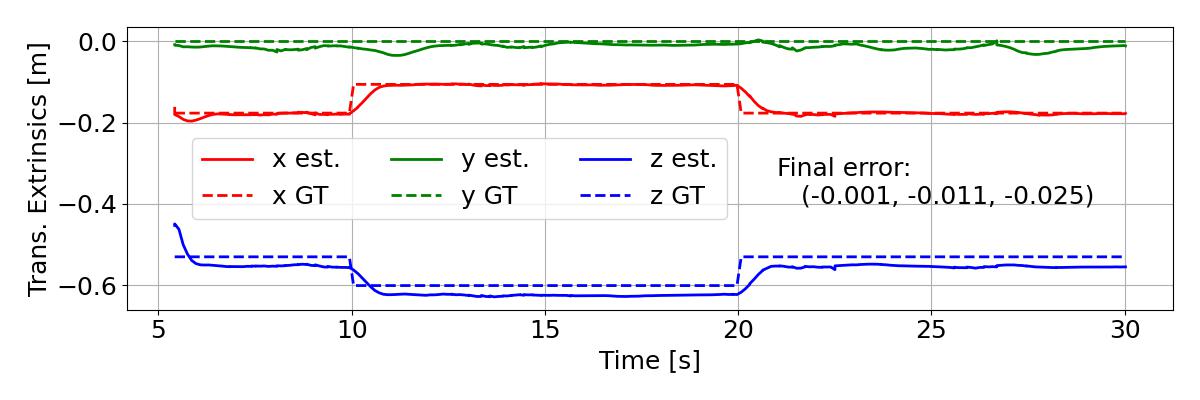}
    \includegraphics[width=\linewidth]{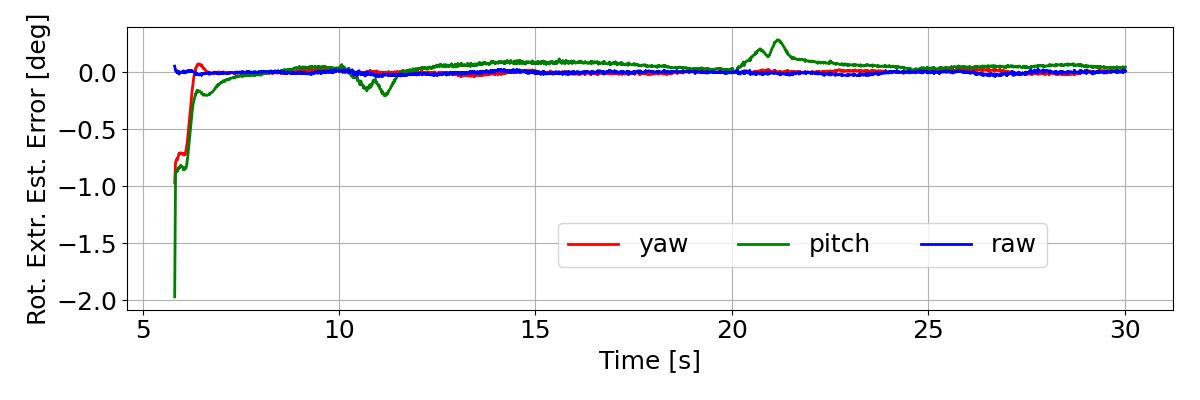}
\caption{The estimation error of the extrinsic parameters between LiDAR 0 and LiDAR 1.}
\label{fig: extrinsic estimation}
\end{figure}

From Fig. \ref{fig: simulation setup}, Fig. \ref{fig: extrinsic estimation}, we can see that the MLCME scheme can estimate the LiDAR extrinsic with the highest accuracy. The extrinsic parameters converge quickly within the first few seconds, and at $t=10$, when there is a significant change in the extrinsic parameters, the MLCME scheme can quickly correct its estimate.
Table \ref{tab: mlcme ape} also shows that our approach achieves the lowest RMSE. Moreover, we can also see that while other methods have their RMSE increase significantly as $\Omega$ increases, the increase of GPTR method is much milder, showing that the third-order model has a higher estimation capacity than the models of lower order.

\begin{table}

    \centering
    \renewcommand{\arraystretch}{1.5}
    \begin{threeparttable}
    \caption{Comparing the RMSE of trajectory estimate by GPTR's MLCME scheme and other methods with the synthetic data at different values of $\bs{\Omega}$.} \label{tab: mlcme ape}
    \begin{tabular}{ccccccc}
        \hline\hline
        \mr{2}{*}{$\bs{\Omega}$}
        &\mr{2}{*}{\bf{Lidar}}
        &\mr{2}{*}{\bf{I$^2$EKF}}
        &\mr{2}{*}{\bf{Traj}}
        &\mr{2}{*}{\bf{CTEM}}
        &\mc{2}{c}{\bf{GPTR}}
        \\\cline{6-7}
        & & & & & {\scriptsize \sepose} & {\scriptsize \srpose} \\\cline{6-7}
        \hline
        \mr[c]{2}{*}{0.25} & 0 & 0.0571 & 0.0635 & 0.0363 & \ul{0.0344} & \bf{0.0343} \\
         & 1 & 0.0495 & 0.0585 & 0.0397 & \bf{0.0392} & \ul{0.0392} \\
        \cline{1-7}
        \mr[c]{2}{*}{0.35} & 0 & 0.0576 & 0.0633 & 0.0359 & \ul{0.0343} & \bf{0.0341} \\
         & 1 & 0.0506 & 0.0586 & 0.0419 & \bf{0.0405} & \ul{0.0405} \\
        \cline{1-7}
        \mr[c]{2}{*}{0.45} & 0 & 0.058 & 0.0632 & 0.0378 & \ul{0.0337} & \bf{0.0332} \\
         & 1 & 0.0531 & 0.0593 & 0.0456 & \bf{0.041} & \ul{0.0412} \\
        \cline{1-7}
        \mr[c]{2}{*}{0.55} & 0 & 0.0583 & 0.0626 & 0.038 & \ul{0.0352} & \bf{0.0347} \\
         & 1 & 0.0564 & 0.0624 & 0.0469 & \bf{0.0427} & \ul{0.0432} \\
        \cline{1-7}
        \mr[c]{2}{*}{0.65} & 0 & 0.0595 & 0.063 & 0.0396 & \ul{0.0348} & \bf{0.0342} \\
         & 1 & 0.0621 & 0.0657 & 0.05 & \bf{0.0436} & \ul{0.0443} \\
        \cline{1-7}
        \mr[c]{2}{*}{0.75} & 0 & 0.0617 & 0.0623 & 0.0426 & \ul{0.0351} & \bf{0.034} \\
         & 1 & 0.0693 & 0.0715 & 0.053 & \bf{0.0443} & \ul{0.0455} \\
        \cline{1-7}
        \mr[c]{2}{*}{0.85} & 0 & 0.0632 & 0.0629 & 0.0456 & \ul{0.0373} & \bf{0.0359} \\
         & 1 & 0.0756 & 0.08 & 0.0556 & \bf{0.0457} & \ul{0.0468} \\
        \cline{1-7}
        \mr[c]{2}{*}{0.95} & 0 & 0.0671 & 0.0641 & 0.0561 & \ul{0.0385} & \bf{0.0374} \\
         & 1 & 0.0856 & 0.0905 & 0.0669 & \bf{0.0475} & \ul{0.0489} \\
        \hline\hline
    \end{tabular}
    \begin{tablenotes}
        \small
        \item Note: All values are in [m] unit. \textbf{Traj} and \textbf{CTEM} are shorthand for Traj-LO \cite{zheng2024traj} and CTE-MLO \cite{shen2024cte}. The GPTR methods use the closed-form kinematic model.
    \end{tablenotes}
    \end{threeparttable}
\end{table}

\begin{remark}
Our example above demonstrates the efficacy of GPTR in a multi-trajectory calibration scenario; thus, a map of the room was assumed available (in practice, it can be obtained by using a 3D scanner or by stitching together some static LiDAR scans). Without a prior map, one can adopt a batch optimization scheme with iterative feature matching \cite{liu2022targetless, li2024pss, li2021towards}. The example can also be modified to become a full-fledged LOAM system with online calibration by integrating some incremental mapping technique, for e.g., ikd-Tree \cite{cai2021ikd}, VoxelMap \cite{yuan2022efficient}, UFOMap \cite{duberg2020ufomap}. To the best of our knowledge, few works have addressed the online calibration of multi-LiDAR systems, and most works often assume a constant LiDAR extrinsic. In the rare case of M-LIOM \cite{jiao2021robust} that does consider online calibration, the method requires overlap between sensors for it to work.
\end{remark}





\section{Conclusion}  \label{sec: conclusion}

In this paper, we have developed a CT trajectory representation framework based on third-order GP. The framework features a closed-form kinematic model for both \srpose\ and \sepose\ pose representation. These kinematics models are encapsulated in a unified framework, thus making it easier for users to experiment and compare different configurations. We also include a comprehensive set of working examples of motion estimation, leveraging LiDAR, visual, IMU, and UWB factors in both batch-optimization and sliding-window optimization schemes. The experiments have also provided crucial insights into the benefits of closed-form kinematic models over traditional approximate models. Based on these examples, we also hope the community can come up with more advanced motion estimation schemes featuring other sensing models. Our code base named GPTR has been open-sourced at \url{https://github.com/brytsknguyen/gptr}, and we welcome all feedback and contributions from the community to improve our project.

\appendices

\section{Closed forms of \texorpdfstring{$\Jr(\bsu)$}{Jr}, \texorpdfstring{$\JrInv(\bsu)$}{JrInv}, \texorpdfstring{$\SOH_1(\bsu, \bsv)$}{Hthe}, \texorpdfstring{$\SOHp_1(\bsu, \bsv)$}{H'the}} \label{app: Jr JrInv H Hprime}

{
\def\U{\bsu}
\def\Un{u}
\def\Ub{\bsub}
\def\V{\bsv}
\def\W{\bsw}
\newcommand{\Jcbg}[2]{\Jcb{g_{#1}^{(#2)}}{\U}}
Let us first define the $f(\cdot)$ functions in \eqref{eq: gf form}:
\begin{align} \label{eq: the f functions}
    \f(\U)
    &\triangleq (\U^\wedge)^2,
    \\
    \f_u(\U, \V)
    &\triangleq \partder{f(\U)\V}{\U} = - \U^\wedge \V^\wedge - (\U^\wedge \V)^\wedge, \label{eq: fu()}
    \\ 
    \f_{uu}(\U, \V, \bsw)
    &\triangleq \partder{\f_u(\U,\V)\W}{\U} = (\V^\wedge \bsw)^\wedge - \bsw^\wedge \V^\wedge,
    \\
    \f_{uv}(\U, \V, \bsw)
    &\triangleq \partder{\f_u(\U,\V)\W}{\V} = \U^\wedge \bsw^\wedge + \bsw^\wedge \U^\wedge,
\end{align}
where $\U, \V, \bsw$ are $\R^3$ vectors.

For a vector $\U$, let us define the following
\begin{align}
    \Un \triangleq \norm{\U},
    \
    \Ub \triangleq \frac{\U}{\Un},
    \
    \Ub_\U \triangleq (\textbf{I} - \Ub\Ub^\top)\frac{1}{\Un}.
\end{align}

We then define the $g(\cdot)$ functions and its derivatives:
\begin{align}
    g_1^{(0)}(u) &= g_1(u)
    \triangleq \frac{1-\cos u}{\usqu},\  
    \label{eq: g10}
    \\
    g_1^{(1)}(u)
    &
     =  \frac{\sin u}{\usqu}
      + \frac{2(\cos u - 1)}{\ucub},
    \\
    g_1^{(2)}(u)
    &
     = \frac{\cosu}{\usqu} - \frac{4\sinu}{\ucub} 
    + \frac{6 - 6\cosu}{\utes};
    \\
    g_2^{(0)}(u) &= g_2(u)
    \triangleq \frac{ u - \sin u}{\ucub},\ 
    \\
    g_2^{(1)}(u)
    &
     = - \frac{2}{\ucub}
       - \frac{\cos u}{\ucub}
       + \frac{3\sin u}{\utes},
    \\
    g_2^{(2)}(u)
    &
     = \frac{\sinu}{\ucub} + \frac{6\cosu + 6}{\utes}
    \quad- \frac{12\sinu}{\uqui};
    \\
    g_3^{(0)}(u) &= g_3(u)
    \triangleq \frac{1}{\usqu} - \frac{1 + \cos u}{2\sin u},
    \\
    g_3^{(1)}(u)
    &
     = \frac{-2}{\ucub} + \frac{(1 + \cosu)(u+\sinu)}{2\usqu(\sinu)^2},
    \\
    g_3^{(2)}(u)
    &
     = \frac{6}{\utes} + \frac{\sinu}{\ucub(\cos u - 1)}
    \nonumber\\
    &\quad
    +\frac{u\cosu + 2\sinu + u}{2\usqu\sinu(\cosu - 1)}.
    \label{eq: g32}
\end{align}

In addition, we also define the following Jacobians
\begin{align}
    &\Jcb{g_j^{(i)}}{\U} \triangleq g_j^{(i+1)}(u)\Ub^\top,
\end{align}

Hence, the closed forms of $\Jr$ and $\JrInv$ as provided in \cite{sola2018micro} are recalled as follows:
\begin{align}
    &\Jr(\U)
    =
    \textbf{I} 
    - g_1(\Un) \U^\wedge
    + g_2(\Un)\f(\U);
    \\    
    &\JrInv(\U)
    =
    \textbf{I} 
    + \frac{1}{2} \U^\wedge
    + g_3(\Un)\f(\U).
\end{align}
Based on the forms of $\Jr(\U)$ and $\JrInv(\U)$ above, we apply the procedure described in Sec. \ref{sec: procedure} to achieve the closed forms of $\SOH_1(\U, \V)$ and $\SOHp_1(\U, \V)$. Specifically,
\begin{align}
    \SOH_1(\U, \V)
    &=
    \V^\wedge g_1(\Un)
    + \V^\wedge \U \Jcbg{1}{0}
    \nonumber
    \\
    &\qquad
    + \f_u(\U, \V) g_2(\Un)
    + \f(\U)\V \Jcbg{2}{0},
    \\
    \SOHp_1(\U, \V)
    &= -\frac{1}{2}\V^\wedge + \f_u(\U, \V)g_3(\Un) + \f(\U)\V\Jcbg{3}{0}.
\end{align}

In Appendix \ref{sec: L mappings}, we apply the procedure in Sec. \ref{sec: procedure} once more on $\SOH$ and $\SOHp$, which will be used to calculate the intrinsic Jacobians in Sec. \ref{sec: instrinsic jacobians}.
}

\section{Closed forms of \texorpdfstring{$\SOL$}{L} and \texorpdfstring{$\SOLp$}{L'} mappings} \label{sec: L mappings}


{

\def\U{\bsu}
\def\Un{u}
\def\Ub{\bsub}

\def\V{\bsv}
\def\W{\bsw}

\newcommand{\Jcbg}[2]{\Jcb{g_{#1}^{(#2)}}{\U}}

\def\DubDu{\partder{\Ub}{\U}}
\def\UbtpW{\bsub^\top\bsw}
\def\WtpDUb{\W^\top\DubDuCP}
\def\Vsk{\bsv^\wedge}

\def\UsksqV{f(\U)\V}
\def\DXsksqVDX{f_u(\U,\V)}
\def\DXsksqVDV{f_v(\U,\V)}
\def\DDXsksqVADXDX{f_{uu}(\U, \V, \W)}
\def\DDXsksqVADXDV{f_{uv}(\U, \V, \W)}


\begin{align}
\begin{split}
    &\SOL_{11}(\U, \V, \W) =\partder{\SOH_1(\U, \V)\W}{\U}
    \\
    &=
     \Vsk\W\Jcbg{1}{0} + \Vsk(\Jcbg{1}{0}\W)
    \\
    &\ 
      + \Vsk\U(\Ub^\top\W)\Jcbg{1}{1} + \Vsk \U g_1^{(1)}(u) \W^\top\Ub_\U
    \\
    &\ 
     +\f_{uu}(\U, \V, \W) g_2(u)
     +\f_{u}(\U, \V)\W \Jcbg{2}{0}
    \\
    &\ 
     +\f_u(\U, \V)(\Jcbg{2}{0}\W)
     +\f(\U,\V)g_2^{(1)}(u)\W^\top\Ub_\U
    \\
    &\ 
     +\f(\U)\V\Ub^\top\W \Jcbg{2}{1},
\end{split}
\end{align}
\begin{align}
\begin{split}
    &\SOL_{12}(\U, \V, \W)
     =\partder{\SOH_1(\U, \V)\W}{\V}
    \\
    &=
     -\W^\wedge g_1(u) - \U^\wedge \Jcbg{1}{0}\W + \f_{uv}(\U, \V, \W)g_2(u)
    \\
    &\quad
     + \f_v(\U)\V(\Jcbg{2}{0}\W),
\end{split}
\end{align}
}

{
\def\U{\bsu}
\def\Un{u}
\def\Ub{\bsub}

\def\V{\bsv}
\def\W{\bsw}

\newcommand{\Jcbg}[2]{\Jcb{g_{#1}^{(#2)}}{\U}}

\def\Xn{\Then}
\def\Xb{\Theb}

\def\DubDu{\partder{\Xb}{\X}}
\def\XbtpA{\Theb^\top\Thed}
\def\WtpDUb{\A^\top\DubDuCP}
\def\Vsk{\Thed^\wedge}

\def\XsksqV{f(\X,\V)}
\def\DXsksqVDX{f_u(\X,\V)}
\def\DXsksqVDV{f_v(\X,\V)}
\def\DDXsksqVADXDX{f_{uu}(\X, \V, \A)}
\def\DDXsksqVADXDV{f_{uv}(\X, \V, \A)}

\begin{align}
\begin{split}
    &\SOLp_{11}(\U, \V, \W)
     =\partder{\SOHp_1(\U, \V)\W}{\U}
    \\
    &=
    \f_{uu}(\U, \V, \W)g_3(u)
    + \f_u(\U, \V)\W\Jcbg{3}{0}
    \\
    &\quad
     +\f_u(\U,\V)\Jcbg{3}{0}\W
     +\f(\U,\V)g_3^{(1)}(u)\W^\top\Ub_\U
    \\
    &\quad
     +\f(\U,\V)\W^\top\Ub\Jcbg{3}{1},
\end{split}
\end{align}

\begin{align}
\begin{split}
    &\SOLp_{12}(\U, \V, \W)
     =\partder{\SOHp_1(\The, \V)\W}{\V}
    \\
    &=
    \ \frac{1}{2}\W^\wedge + \f_{uv}(\U, \V, \W) g_3^{(0)}(u) + \f_v(\U, \V, \W)\Jcbg{3}{0}.
\end{split}
\end{align}
}

\section{The Intrinsic Jacobians} \label{sec: instrinsic jacobians}

\subsection{\texorpdfstring{\srpose}{SO(3)xR3} Representation} \label{sec: instrinsic jacobians srpose}

Let us first detail the intrinsic Jacobians of the $\SO$ GP. Afterwards, the vector GP can be laid out easily.
To simplify the notation, let us denote $\Jcb{\Y}{\X}\triangleq\partder{\Y}{\X}$. In Fig. \ref{fig: jacobian heirarchy}, we provide a diagram of the relationship between the variables. Our goal is to find all of the Jacobians $\Jcb{\Y}{\X}$ where $\Y$ belongs to L4 and $\X$ belongs to L1. This can be done in a layer-by-layer manner as follows.

First, we calculate the Jacobians of variables in L2 over those in L1. By using the diagram in Fig. \ref{fig: jacobian heirarchy}, we can identify eleven L1-L0 connections, each can be represented by one Jacobian. Note that a connection can have multiple distinct paths. Specifically, from $\bs{\gamma}_a$ states, we have two connections by two simple paths:
\begin{align}
    \Jcb{\Thed_a}{\angvel_a}
    &= \textbf{I}
    ,
    \ 
    \Jcb{\Thedd_a}{\angacc_a}
    = \textbf{I}
    .
\end{align}

From $\The_b$, we can identify two connections:
\begin{align}
    \Jcb{\The_b}{\rot_a}
    &= \partder{\Log(\rot_a^{-1}\rot_b)}{\rot_a}
    = -\JrInv(-\The_b)
    ,
    \\
    \Jcb{\The_b}{\rot_b}
    &= \partder{\Log(\rot_a^{-1}\rot_b)}{\rot_b}
    = \JrInv(\The_b){}
    ,
\end{align}

From $\Thed_b$, there are three connections based on three paths:
\begin{align}
    \Jcb{\Thed_b}{\rot_a}
    &= \partder{[\JrInv(\The_b)\angvel_b]}{\rot_a}
    = \partder{[\JrInv(\The_b)\angvel_b]}{\The_b} \partder{\The_b}{\rot_a} 
    \nonumber\\
    &= \SOHp_1(\The_b, \angvel_b)\Jcb{\The_b}{\rot_a}
    ,
    \\
    \Jcb{\Thed_b}{\rot_b}
    &= \partder{[\JrInv(\The_b)\angvel_b]}{\rot_b}
    = \SOHp_1(\The_b, \angvel_b)\Jcb{\The_b}{\rot_b}
    ,
    \\
    \Jcb{\Thed_b}{\angvel_b}
    &= \JrInv(\The_b)
    ,
\end{align}

From $\Thedd_b$, there are two paths to $\rot_a$, hence:
\begin{align}
    \Jcb{\Thedd_b}{\rot_a}
    &= \partder{[\JrInv(\The_b)\angacc_b + \SOHp_1(\The_b, \angvel_b)\Thed_b]}{\rot_a}
    \nonumber\\
    &= \SOHp_1(\The_b, \angacc_b)\Jcb{\The_b}{\rot_a}
    + \SOHp_1(\The_b, \angvel_b)\Jcb{\Thed_b}{\rot_a}
    \nonumber\\
    &\quad+ \SELp_{11}(\The_b, \angvel_b, \Thed_b)\Jcb{\The_b}{\rot_a},
\end{align}
where $L'$ is defined in Appendix \ref{sec: L mappings}.

Similarly, from $\Thedd_b$ to $\rot_b$, there are three paths, $\Thedd_b$ to $\angvel_b$ two, and $\Thedd_b$ to $\angacc_b$ one, yielding three Jacobians:
\begin{align}
    \Jcb{\Thedd_b}{\rot_b}
    &= \SOHp_1(\The_b, \angacc_b)\Jcb{\The_b}{\rot_b}
    + \SOHp_1(\The_b, \angvel_b)\Jcb{\Thed_b}{\rot_b}
    \nonumber\\
    &\quad + \SELp_{11}(\The_b, \angvel_b, \Thed_b)\Jcb{\The_b}{\rot_b}
    \\
    \Jcb{\Thedd_b}{\angvel_b}
    &= \SELp_{12}(\The_b, \angvel_b, \Thed_b) + \SOHp_1(\The_b, \angvel_b)\Jcb{\Thed_b}{\angvel_b}
    \\
    \Jcb{\Thedd_b}{\angacc_b}
    &= \JrInv(\The_b)
\end{align}

Next for the Jacobians between L3 and L2, from \eqref{eq: interpolation} we have the following interpolation relationship:
\begin{align}
\begin{bmatrix}
    \Thet
    \\
    \Thedt
    \\
    \Theddt
\end{bmatrix}
=
\begin{sqbmat}
    \ddots \\
    &\bs{\Lambda}_{ij}\\
    &&\ddots
\end{sqbmat}
\begin{bmatrix}
    \The_a
    \\
    \Thed_a
    \\
    \Thedd_a
\end{bmatrix}
+
\begin{sqbmat}
    \ddots \\
    &\bs{\Psi}_{ij}\\
    &&\ddots
\end{sqbmat}
\begin{bmatrix}
    \The_b
    \\
    \Thed_b
    \\
    \Thedd_b
\end{bmatrix}
\end{align}
where $\bs{\Lambda}_{ij}, \bs{\Psi}_{ij} \in \R^{3\times3}$ are some $3\times3$ blocks of $\bs{\Lambda}(\tau)$ and $\bs{\Psi}(\tau)$. Thus, the Jacobians from L2 to L1 can be easily calculated as:
\begin{equation} \label{eq: J local variables}
    \Jcb{\Thet^{(i)}}{\The^{(j)}_a} = \bs{\Lambda}_{i,j},\quad \Jcb{\Thet^{(i)}}{\The^{(j)}_b} = \bs{\Psi}_{i,j},\ i, j \in {0, 1, 2},
\end{equation}
where $\The^{(0)} \triangleq \The$, $\The^{(1)} \triangleq \Thed$, and $\The^{(2)} \triangleq \Thedd$.

Finally for the Jacobian from L4 to L3, from the mappings \eqref{eq: local to global mappings}, we have:
\begin{align}
    \Jcb{\rot_t}{\The_t}
    &= \partder{\rot_a \Exp(\Thet)}{\Thet}
     = \Jr(\Thet),
    \\
    \Jcb{\angvel_t}{\Thet}
    &= \partder{\Jr(\Thet) \Thedt}{\Thet}
     = \SOH_1(\The_t, \Thed_t),
    \\
    \Jcb{\angvel_t}{\Thedt}
    &= \partder{\Jr(\Thet) \Thedt}{\Thedt}
     = \Jr(\Thet),
    \\
    \Jcb{\angacc_t}{\Thet}
    &= \SOH_1(\Thet, \Theddt) + \SEL_{11}(\Thet, \Thedt, \Thedt),
    \\
    \Jcb{\angacc_t}{\Thedt}
    &= \SOH_1(\Thet, \Thedt) + \SEL_{12}(\Thet, \Thedt, \Thedt),
    \\
    \Jcb{\angacc_t}{\Theddt}
    &= \Jr(\Thet).
\end{align}

Now, we can find the Jacobians from L3 to L1 by ``merging'' the L3-L2 and L2-L1 Jacobians as follows:
\begin{align}
    \Jcb{\Thet^{(i)}}{\rot_a}
    &=
      \Jcb{\Thet^{(i)}}{\The_b}\Jcb{\The_b}{\rot_a}
    + \Jcb{\Thet^{(i)}}{\Thed_b}\Jcb{\Thed_b}{\rot_a}
    + \Jcb{\Thet^{(i)}}{\Thedd_b}\Jcb{\Thedd_b}{\rot_a},
    \\
    \Jcb{\Thet^{(i)}}{\rot_b}
    &=
      \Jcb{\Thet^{(i)}}{\The_b}\Jcb{\The_b}{\rot_b}
    + \Jcb{\Thet^{(i)}}{\Thed_b}\Jcb{\Thed_b}{\rot_b}
    + \Jcb{\Thet^{(i)}}{\Thedd_b}\Jcb{\Thedd_b}{\rot_b},
    \\
    \Jcb{\Thet^{(i)}}{\angvel_a}
    &=
      \Jcb{\Thet^{(i)}}{\Thed_a}\Jcb{\Thed_a}{\angvel_a},
    \ 
    \Jcb{\Thet^{(i)}}{\angvel_b}
    =
      \Jcb{\Thet^{(i)}}{\Thed_b}\Jcb{\Thed_b}{\angvel_b}
     +\Jcb{\Thet^{(i)}}{\Thedd_b}\Jcb{\Thedd_b}{\angvel_b},
    \\
    \Jcb{\Thet^{(i)}}{\angacc_a}
    &=
      \Jcb{\Thet^{(i)}}{\Thedd_a}\Jcb{\Thedd_a}{\angacc_a},
    \ 
    \Jcb{\Thet^{(i)}}{\angacc_b}
    =
      \Jcb{\Thet^{(i)}}{\Thedd_b}\Jcb{\Thedd_b}{\angacc_b},\ i \in \{0, 1, 2\}.
\end{align}

Moving on to L4, we note that there is one direct path from L4 to L1, which corresponds to the Jacobian:
\begin{equation} \label{eq: l3-l0}
    \partder{\rot_a \Exp(\phi)}{\rot_a}\vert_{\phi=\Thet}
    = \vbf{Ad}_{\Exp(\Thet)}^{-1} = \Exp(-\Thet).
\end{equation}
Finally, we can assemble the L4-L1 Jacobians from \eqref{eq: l3-l0} and the L4-L3 and L3-L1 Jacobians above. For $\rot_t$, we have:
\begin{align}
    \Jcb{\rot_t}{\rot_a}
    &= \Exp(-\Thet) + \Jcb{\rot_t}{\Thet}\Jcb{\Thet}{\rot_a}
    ,
    \ 
    \Jcb{\rot_t}{\rot_b}
    =
    \Jcb{\rot_t}{\Thet}\Jcb{\Thet}{\rot_b},
    \\
    \Jcb{\rot_t}{\angvel_a}
    &= \Jcb{\rot_t}{\Thet}\Jcb{\Thet}{\angacc_a}
    ,\ 
    \Jcb{\rot_t}{\angvel_b}
    = \Jcb{\rot_t}{\Thet}\Jcb{\Thet}{\angvel_b}
    ,\\
    \Jcb{\rot_t}{\angacc_a}
    &= \Jcb{\rot_t}{\Thet}
          \Jcb{\Thet}{\angacc_a}
    ,\ 
    \Jcb{\rot_t}{\angacc_b}
    = \Jcb{\rot_t}{\Thet}\Jcb{\Thet}{\angacc_b},
\end{align}
and similarly, for $\angvel_t$ and $\angacc_t$:
\begin{align}
    \Jcb{\angvel_t}{\rot_j}
    &=
    \Jcb{\angvel_t}{\Thet}
    \Jcb{\Thet}{\rot_j}
    +
    \Jcb{\angvel_t}{\Thedt}
    \Jcb{\Thedt}{\rot_j}
    ,
    \\
    \Jcb{\angvel_t}{\angvel_j}
    &=
    \Jcb{\angvel_t}{\Thet}
    \Jcb{\Thet}{\angvel_j}
    +
    \Jcb{\angvel_t}{\Thedt}
    \Jcb{\Thedt}{\angvel_j}
    ,
    \\
    \Jcb{\angvel_t}{\angacc_j}
    &=
    \Jcb{\angvel_t}{\Thet}
    \Jcb{\Thet}{\angacc_j}
    +
    \Jcb{\angvel_t}{\Thedt}
    \Jcb{\Thedt}{\angacc_j}
    ,
    \\
    \Jcb{\angacc_t}{\rot_i}
    &=
    \Jcb{\angacc_t}{\Thet}
    \Jcb{\Thet}{\rot_i}
    +
    \Jcb{\angacc_t}{\Thedt}
    \Jcb{\Thedt}{\rot_i}
    +
    \Jcb{\angacc_t}{\Theddt}
    \Jcb{\Theddt}{\rot_i}
    ,
    \\
    \Jcb{\angacc_t}{\angvel_i}
    &=
    \Jcb{\angacc_t}{\Thet}
    \Jcb{\Thet}{\angvel_i}
    +
    \Jcb{\angacc_t}{\Thedt}
    \Jcb{\Thedt}{\angvel_i}
    +
    \Jcb{\angacc_t}{\Theddt}
    \Jcb{\Theddt}{\angvel_i},
    \\
    \Jcb{\angacc_t}{\angacc_i}
    &=
    \Jcb{\angacc_t}{\Thet}
    \Jcb{\Thet}{\angacc_i}
    +
    \Jcb{\angacc_t}{\Thedt}
    \Jcb{\Thedt}{\angacc_i}
    +
    \Jcb{\angacc_t}{\Theddt}
    \Jcb{\Theddt}{\angacc_i}
    ,
    \ 
    i \in \{a, b\}.
\end{align}

To be complete, the Jacobian of the interpolated $\R^3$ states $\pos_t$, $\vel_t$, $\acc_t$ over $\pos_a$, $\vel_a$, $\acc_a$ and $\pos_b$, $\vel_b$, $\acc_b$ resemble \eqref{eq: J local variables}:
\begin{equation}
    \Jcb{\pos_t^{(i)}}{\pos^{(j)}_a} = \bs{\Lambda}_{i,j},\quad \Jcb{\pos_t^{(i)}}{\pos^{(j)}_b} = \bs{\Psi}_{i,j},\ i, j \in {0, 1, 2},
\end{equation}
where $\pos^{(0)} = \pos$, $\pos^{(1)} = \vel$, $\pos^{(2)} = \acc$.

\subsection{\texorpdfstring{$\SE$}{SE(3)} Representation} \label{sec: instrinsic jacobians sepose}

As can be seen in Fig. \ref{fig: jacobian heirarchy se3}, the connections from layers L4 to L1 mirror the $\SO$ case.
Hence, the Jacobians from variables of L4 \wrt L1 can be directly adapted from those in Sec. \ref{sec: gptr on SO3}. Specifically, to obtain these Jacobians, the readers can respectively replace the variables $\The$, $\Thed$, $\Thedd$, $\rot$, $\angvel$, $\angacc$, $\Jr$, $\JrInv$, $H$, $H'$, $L$, $L'$ in the $\SO$ case by $\Xi$, $\Xid$, $\Xidd$, $\tf$, $\twist$, $\wrench$, $\SEJr$, $\SEJrInv$, $\SEH$, $\SEHp$, $\SEL$, $\SELp$, whereas $\SEL$, $\SELp$ are defined as:
\begin{align}
    &\SEL_{11}(\Xi, \Xid, \Xid)
    =
    \mypartder{\SEH_1(\Xi, \Xid)\Xid}{\Xi}
    \nonumber
    \\
    &=
    \begin{bmatrix}
        \mypartder{\SOH_1(\The, \Thed)\Thed}{\The}
        &0
        \\
        \mypartder{
        \begin{aligned}
            &\SES_1(\Xi, \Thed)\Thed
            \\&+\SOH_1(\The, \bsrhod)\Thed
            \\&+\SES_2(\Xi, \Thed)\bsrhod
        \end{aligned}
        }{\The}
        &\mypartder{
        \begin{aligned}
            &\SES_1(\Xi, \Thed)\Thed
            \\&+\SES_2(\Xi, \Thed)\bsrhod
        \end{aligned}
        }{\bsrho}
    \end{bmatrix}
    \nonumber
    \\
    &=
    \begin{bmatrix}
        \SEL_{11}(\The, \Thed, \Thed)
        &0
        \\
        \ll[
        \begin{aligned}
            &\SEC_{11}(\Xi, \Thed, \Thed)
            \\&+\SEL_{11}(\The, \bsrhod, \Thed)
            \\&+\SEC_{21}(\Xi, \Thed, \bsrhod)
        \end{aligned}
        \rr]
        &
        \ll[
        \begin{aligned}
            &\SEC_{12}(\Xi, \Thed, \Thed)
            \\&+\SEC_{22}(\Xi, \Thed, \bsrhod)
        \end{aligned}
        \rr],
    \end{bmatrix}
    \end{align}
    \begin{align}
    &\SEL_{12}(\Xi, \Xid, \Xid)
    =
    \mypartder{\SEH_1(\Xi, \Xid)\bsW}{\Xid}
    \bigg|_{\bsW = (\bsW_1, \bsW_2) = \Xid}
    \nonumber
    \\
    &=
    \begin{bmatrix}
        \mypartder{\SOH_1(\The, \Thed)\bsW_1}{\Thed}
        &0
        \\
        \mypartder{
        \begin{aligned}
            &\SES_1(\Xi, \Thed)\bsW_1
            \\&+\SES_2(\Xi, \Thed)\bsW_2
        \end{aligned}}{\Thed}
        &\mypartder{\SOH_1(\The, \bsrhod)\Thed}{\bsrhod}
    \end{bmatrix}
    \bigg|_{\bsW = \Xid}
    \nonumber
    \\
    &=
    \begin{bmatrix}
        L_{12}(\The, \Thed, \Thed)
        &0
        \\
        \SEC_{13}(\Xi, \Thed, \Thed) + \SEC_{23}(\Xi, \Thed, \bsrhod)
       &L_{12}(\The, \bsrhod, \Thed).
    \end{bmatrix}
\end{align}
\begin{align}
    &\SEL'_{11}(\Xi, \twist, \Xid)
    =
    \mypartder{\SEHp_1(\Xi, \twist)\Xid}{\Xi}
    \nonumber
    \\
    &\quad=
    \begin{bmatrix}
        \mypartder{\SOHp_1(\The, \angvel)\Thed}{\The}
        &0
        \\
        \mypartder{
        \begin{aligned}
            &\SESp_1(\Xi, \angvel)\Thed
            \\&+\SOHp_1(\The, \bsnu)\Thed
            \\&+\SESp_2(\Xi, \angvel)\bsrhod
        \end{aligned}
        }{\The}
        &\mypartder{
        \begin{aligned}
            &\SESp_1(\Xi, \angvel)\Thed
            \\&+\SESp_2(\Xi, \angvel)\bsrhod
        \end{aligned}
        }{\bsrho}
    \end{bmatrix}
    \nonumber
    \\
    &\quad=
    \begin{bmatrix}
        \SELp_{11}(\The, \angvel, \Thed)
        &0
        \\
        \ll[
        \begin{aligned}
            &\SECp_{11}(\Xi, \angvel, \Thed)
            \\&+\SELp_{11}(\The, \bsnu, \Thed)
            \\&+\SECp_{21}(\Xi, \angvel, \bsrhod)    
        \end{aligned}
        \rr]
        &
        \ll[
        \begin{aligned}
            &\SECp_{12}(\Xi, \angvel, \Thed)
            \\&+\SECp_{22}(\Xi, \angvel, \bsrhod)
        \end{aligned}
        \rr]
    \end{bmatrix}
    \end{align}
    \begin{align}
    &\SEL_{12}(\Xi, \twist, \Xid)
    =
    \mypartder{\SEHp_1(\Xi, \twist)\Xid}{\twist}
    \nonumber
    \\
    &=
    \begin{bmatrix}
        \mypartder{\SOHp_1(\The, \angvel)\Thed}{\angvel}
        &0
        \\
        \mypartder{
        \begin{aligned}
            &\SESp_1(\Xi, \angvel)\Thed
            \\&+\SESp_2(\Xi, \angvel)\bsrhod
        \end{aligned}
        }{\angvel}
        &\mypartder{\SOHp_1(\The, \bsnu)\Thed}{\bsnu}
    \end{bmatrix}
    \nonumber
    \\
    &=
    \begin{bmatrix}
        \SELp_{12}(\The, \angvel, \Thed)
        &0
        \\
        \SECp_{13}(\Xi, \angvel, \Thed) + \SECp_{23}(\Xi, \angvel, \bsrhod)
       &\SELp_{12}(\The, \bsnu, \Thed)
    \end{bmatrix}
\end{align}
The matrices $\SES$, $\SEC$, $\SESp$ and $\SECp$ are calculated using the procedure in Sec. \ref{sec: procedure}.

Then we find the Jacobians from layer L5 to L4 based on the following mappings:
\begin{align}
    \rot_t &= \textbf{M}\tf_t\textbf{M}^\top,
    \ \vbf{M} = [\vbf{I}_{3\times3}\ \bzr_{3\times1}], \label{eq: R wrt T}
    \\
    \pos_t &= \textbf{M}\tf_t\textbf{N},
    \ \vbf{N} = [\bzr_{3\times3},\ \vbf{1}_{3\times1}]^\top,
    \nonumber\\
    \angvel_t &= \textbf{U}\twist_t,
    \ \vbf{U} = [\vbf{I}_{3\times3},\ \bzr_{3\times3}],
    \\
    \vel_t &= \rot_t\textbf{D}\twist_t = \rot_t\bsnu_t,
    \ \vbf{D} = [\bzr_{3\times3},\ \vbf{I}_{3\times3}],
    \\
    \angacc_t &= \textbf{U}\wrench_t,
    \\
    \acc_t &= \rot_t\textbf{D}\wrench_t + \rot_t\angvel_t^\wedge\textbf{D}\twist_t = \rot_t\bsbeta_t + \rot_t\angvel_t^\wedge\bsnu_t, \label{eq: a wrt wrench}
\end{align}
which provides the following Jacobians between L5 and L4
\begin{align}
    &\Jcb{\rot_t}{\tf_t} = \vbf{U},
    \ 
    \Jcb{\angvel_t}{\twist_t} = \vbf{U},
    \ 
    \Jcb{\angacc_t}{\wrench_t} = \vbf{U},
    \nonumber\\
    &\Jcb{\pos_t}{\tf_t} = \rot_t\vbf{D},
    \
    \Jcb{\vel_t}{\twist_t} = \rot_t\vbf{D},
    \
    \Jcb{\acc_t}{\wrench_t} = \rot_t\vbf{D},
    \\
    &\Jcb{\vel_t}{\rot_t} = -\rot_t\bsnu_t^\wedge
    \to
    \Jcb{\vel_t}{\tf_t} = \Jcb{\vel_t}{\rot_t}\Jcb{\rot_t}{\tf_t},
    \nonumber\\
    &\Jcb{\acc_t}{\rot_t} = -\rot_t\bsbeta_t^\wedge - \rot_t(\angvel_t^\wedge\bsnu_t)^\wedge
    \to
    \Jcb{\acc_t}{\tf_t} = \Jcb{\acc_t}{\rot_t}\Jcb{\rot_t}{\tf_t},
    \nonumber\\
    &\Jcb{\acc_t}{\angvel_t} = \Jcb{\vel_t}{\rot_t},
    \to
    \Jcb{\acc_t}{\twist_t} = \Jcb{\acc_t}{\angvel_t}\Jcb{\angvel_t}{\twist_t} +\rot_t\angvel_t^\wedge\vbf{D},
\end{align}

Similarly, between L1 and L0, we have the mappings:
\begin{align} \label{eq: T twist wrench wrt to rospva}
    \tf_i
    &=
    \begin{bmatrix}
        \rot_i  &\pos_i\\
        \bzr &1
    \end{bmatrix}
    ,
    \
    \twist_i
    = 
    \begin{bmatrix}
        \angvel_i
        \\
        \rot_i^\top\vel_i
    \end{bmatrix},
    \nonumber\\
    \wrench_i
    &= 
    \begin{bmatrix}
        \angacc_i
        \\
        \rot_i^\top \acc_i - \angvel_i^\wedge\rot_i^\top\vel_i
    \end{bmatrix},
\end{align}
which provides the following Jacobians
\begin{align}
    &\Jcb{\tf_i}{\rot_i} = \vbf{U}^\top,\ \Jcb{\tf_i}{\pos_i} = \vbf{D}^\top\rot_i^\top,
    \\
    &\Jcb{\twist_i}{\angvel_i} = \vbf{U}^\top,
    \ 
    \Jcb{\twist_i}{\rot_i} = \vbf{D}^\top(\rot_i^\top\vel_i)^\wedge,
    \ 
    \Jcb{\twist_i}{\vel_i} = \vbf{D}^\top\rot_i^{\top},
    \\
    &\Jcb{\wrench_i}{\angacc_i} = \vbf{U}^\top,
    \ 
    \Jcb{\wrench_i}{\rot_i} = \vbf{D}^\top\ll[(\rot_i^\top\acc_i)^\wedge - \angvel_i^\wedge(\rot_i^\top\vel_i)^\wedge\rr],
    \\ 
    &
    \Jcb{\wrench_i}{\acc_i} = \vbf{D}^\top\rot_i^\top,
    \ 
    \Jcb{\wrench_i}{\angvel_i} = \Jcb{\twist_i}{\rot_i},
    \ 
    \Jcb{\wrench_i}{\vel_i} = -\vbf{D}^\top \angvel_i^\wedge\rot_i^\top.
    \label{eq: L1 L0 Jcb}
\end{align}

Hence, we can combine L4-L1 and L1-L0 Jacobians to obtain L4-L0 Jacobians. Specifically for $i \in \{a, b\}$, we have:
\begin{align}
    &\Jcb{\X_t}{\rot_i}
    = \Jcb{\X_t}{\tf_i}\Jcb{\tf_i}{\rot_i}
    + \Jcb{\X_t}{\twist_i}\Jcb{\twist_i}{\rot_i}
    + \Jcb{\X_t}{\wrench_i}\Jcb{\wrench_i}{\rot_i},
    \ 
    \Jcb{\X_t}{\pos_i}
    = \Jcb{\X_t}{\tf_i}\Jcb{\tf_i}{\pos_i}
    \\
    &\Jcb{\X_t}{\angvel_i}
    = \Jcb{\X_t}{\twist_i}\Jcb{\twist_i}{\angvel_i}
    + \Jcb{\X_t}{\wrench_i}\Jcb{\wrench_i}{\angvel_i},
    \ 
    \Jcb{\X_t}{\vel_i}
    = \Jcb{\X_t}{\twist_i}\Jcb{\twist_i}{\vel_i}
    + \Jcb{\X_t}{\wrench_i}\Jcb{\wrench_i}{\vel_i},
    \\
    &\Jcb{\X_t}{\angacc_i}
    = \Jcb{\X_t}{\wrench_i}\Jcb{\wrench_i}{\angacc_i},
    \ 
    \Jcb{\X_t}{\acc_i}
    = \Jcb{\X_t}{\wrench_i}\Jcb{\wrench_i}{\acc_i}, \X _t \in \{\tf_t, \twist_t, \wrench_t\}.
\end{align}

Finally, we can prepend the L5-L4 Jacobians to the L4-L0 to obtain the L5-L0 Jacobians. For $\Y_i \in$ L0 we have
\begin{align}
    &\Jcb{\X_t}{\Y_i}
    = \Jcb{\X_t}{\tf_t}\Jcb{\tf_t}{\Y_i},\ \X_i \in \{\rot_i, \pos_i\},
    \\
    &\Jcb{\angvel_t}{\Y_i}
    = \Jcb{\angvel_t}{\tf_t}\Jcb{\tf_t}{\Y_i},
    \ 
    \Jcb{\angacc_t}{\Y_i}
    = \Jcb{\angacc_t}{\tf_t}\Jcb{\tf_t}{\Y_i},
    \\
    &\Jcb{\vel_t}{\Y_i}
    = \Jcb{\vel_t}{\tf_t}\Jcb{\tf_t}{\Y_i}
    + \Jcb{\vel_t}{\twist_t}\Jcb{\twist_t}{\Y_i},
    \\
    &\Jcb{\acc_t}{\Y_i}
    = \Jcb{\acc_t}{\tf_t}\Jcb{\tf_t}{\Y_i}
    + \Jcb{\acc_t}{\twist_t}\Jcb{\twist_t}{\Y_i}
    + \Jcb{\acc_t}{\wrench_t}\Jcb{\wrench_t}{\Y_i}.
\end{align}

\balance
\bibliographystyle{IEEEtran}
\bibliography{references,bibExternal}

\end{document}